\def\year{2022}\relax
\documentclass[letterpaper]{article} 
\usepackage{aaai22}  
\usepackage{times}  
\usepackage{helvet}  
\usepackage{courier}  
\usepackage[hyphens]{url}  
\usepackage{graphicx} 
\usepackage{natbib}  
\usepackage{caption} 
\DeclareCaptionStyle{ruled}{labelfont=normalfont,labelsep=colon,strut=off} 
\usepackage{algorithm}
\usepackage{algorithmic}
\usepackage{xcolor}
\usepackage{amsmath}
\allowdisplaybreaks

\usepackage{newfloat}
\usepackage{listings}
\usepackage{multirow}
\usepackage{subfigure}
\usepackage{amssymb}
\usepackage{amsmath}
\usepackage{pifont}

\newcommand{\xmark}{\ding{55}}
\usepackage{bm}
\usepackage[colorlinks,
linkcolor=black,
anchorcolor=blue,
citecolor=black
]{hyperref}

\floatstyle{ruled}
\newfloat{listing}{tb}{lst}{}
\floatname{listing}{Listing}
\title{CCTrans: Simplifying and Improving Crowd Counting with Transformer}
\author{Anonymous}
\author{
    Ye Tian$^{1}$\thanks{Work done during the internship at Meituan.}, Xiangxiang Chu$^{2}$, Hongpeng Wang$^{1}$
}
\affiliations{
    \textsuperscript{\rm 1}
    Harbin Institute of Technology (Shenzhen)\\
    \textsuperscript{\rm 2}
    Meituan\\
    \textsuperscript{\rm 1}19S151092@stu.hit.edu.cn, wanghp@hit.edu.cn \textsuperscript{\rm 2}chuxiangxiang@meituan.com
   
}

\usepackage{bibentry}

\begin{document}

\maketitle

\begin{abstract}
Most recent methods used for crowd counting are based on the convolutional neural network (CNN), which has a strong ability to extract local features. But CNN inherently fails in modeling the global context due to the limited receptive fields.  However, the transformer can model the global context easily. In this paper, we propose a  simple approach called CCTrans to simplify the design pipeline. Specifically, we utilize a pyramid vision transformer backbone to capture the global crowd information, a pyramid feature aggregation (PFA) model to combine low-level and high-level features,  an efficient regression head with multi-scale dilated convolution (MDC) to predict density maps. Besides, we tailor the loss functions for our pipeline. Without bells and whistles, extensive experiments demonstrate that our method achieves new state-of-the-art results on several benchmarks both in weakly and fully-supervised crowd counting. Moreover, we currently rank No.1 on the leaderboard of NWPU-Crowd. Our code will be made available.  


\end{abstract}

\section{Introduction}
As a research hot topic of computer vision, crowd counting is to estimate the number of crowds in a scene, which is applied in many fields such as urban planning and traffic supervision. Mainstream methods  focus on designing various convolutional neural networks.  

However, there are still two challenges in crowd counting from the generation of images.  Crowds near the camera have larger scales as well as lower densities and vice versa, which causes dramatic scale and density variations within an image. One well-recognized solution is enhancing the global context modeling capability of CNN-based models. But it is not perfect  due to  limited receptive fields. Therefore, researchers propose various mechanisms to refine CNN-based models. A popular approach is designing multi-column architectures with input images in different resolution ratios to extract features of crowds in different scales and densities \cite{zhang2016single,liu2019crowd,sam2017switching,sindagi2017generating}.  But the model structure of these methods is inefficient with many redundant blocks.
 Another approach is introducing auxiliary tasks into crowd counting to capture global semantic information better at the cost of higher complexity and training time \cite{shi2019revisiting,shi2019counting,jiang2020attention,liu2019exploiting}.

   Recent studies focus on designing different attention mechanisms to focus on scale and density variations in global context\cite{yan2019perspective,liu2019recurrent,liu2019context,jiang2020attention,liu2019adcrowdnet}. However, these pipelines are usually complicated with many sensitive hyper-parameters, which require careful tuning for different datasets.
   
   As an orthogonal approach, some work  improves the performance by optimizing novel image augmentation and loss functions \cite{li2018csrnet,yang2020reverse,wang2020distribution,ma2019bayesian}. Nevertheless, it usually needs sufficient data and expert experience. The design is also sophisticated without  significant improvements.
   
Vision transformer has aroused great interest in the  whole community and shows promising results across many vision tasks \cite{dosovitskiy2021an,chu2021twins,touvron2020deit,liu2021swin,chu2021conditional,SETR}. One great advantage of transformer is able to capture long context dependency and enjoy global receptive fields.  Regarding  semantic segmentation, which is a  point-level prediction task like crowd counting, \citet{SETR,chu2021twins,liu2021swin} construct transformer-based models to achieve better performance than CNN baselines. A natural question is whether we can simplify the complicated pipelines of crowd counting by using transformers. The focus of our paper is finding the answer.
\begin{figure*}[h]
	\centering
	\includegraphics[width=\textwidth]{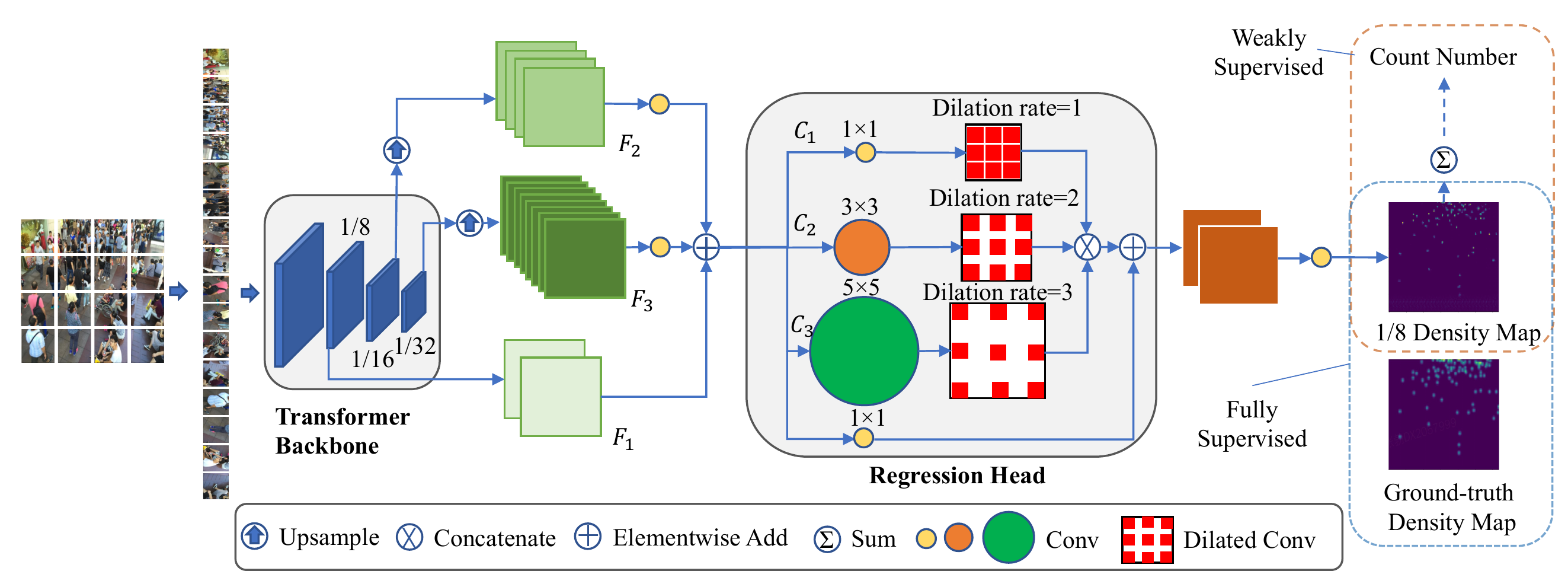}
	\caption{The pipeline of CCTrans. The input image is transformed into a 1D sequence firstly, then the output is fed into the transformer-based backbone. We adopt a pyramid transformer \cite{chu2021twins} to capture global context through various downsampling stages. The outputs of each stage are reshaped into 2D feature maps for pyramid feature aggregation. Finally, a simple regression head with multi-scale receptive fields is used to regress the final results. We support two fashions of supervision. For fully-supervised manner, CCTrans regresses a density map. For weakly-supervised manner, CCTrans sums up all the pixel values of the predicted density map as the crowd number for counting regression. }
	\label{fig:pipeline}
	\centering
\end{figure*}

 In summary, the contributions of this paper are 4-fold:
\begin{itemize}
	\item To simplify the pipelines, we utilize the transformer to construct a simple but high-performance crowd counting model called CCTrans, which can extract semantic features with global context information.
	\item We design an effective feature aggregation block and a simple regression head with multi-scale receptive fields. With two simple designed blocks, we can strengthen the captured features and get accurate regression results.
	\item To further strengthen our method, we tailor the loss functions in both weakly and fully-supervised manners for our method. Specifically, we utilize a smooth weighted loss for the former and smooth $L_1$ for the latter.
	\item Extensive experiments across five popular benchmarks including UCF\_CC\_50, ShanghaiTech Part A and Part B, UCF\_QNRF, and NWPU-Crowd, show that our method achieves new state-of-the-art results under both weakly and fully-supervised settings.
	Moreover, CCTrans ranks No.1 on the leaderboard of NWPU-Crowd.
\end{itemize}

\section{Proposed Method}
The architecture of our method is shown in Figure~\ref{fig:pipeline}. Firstly, the input images are split into fixed-size image patches. Then, the output is flattened to a 1D sequence of vectors. Next, a pyramid transformer backbone is used to extract global features from the sequences. Then the 1D sequences of each stage are reshaped into 2D feature maps and up-sampled to the same resolution. And then, an elementwise addition is performed on these feature maps. Finally, a simple regression head with multi-scale receptive fields is applied to regress the density map. The final density map and the sum of all its pixel values are used to build the loss functions of weakly and fully-supervised manners, respectively. 

\subsection{2D Image to 1D Sequence}
 An image is transformed  into 1D sequences before going into the transformer. We denote an input image as $I \in \mathbb{R}^{H \times W \times 3}$, where $H$, $W$, and 3 are respectively its height, width, and channel size. We then split it into $\frac{H}{K} \times \frac{W}{K}$ image patches and each patch is of size $K\times K\times 3$. 
 We flatten this 2D patch array into a 1D patch sequence $x\in \mathbb{R}^{N \times D}$, $N = \frac{HW}{K^2}$, $D = K\times K\times 3$. 
We use patch embedding for sequence x by applying a learnable projection $f:x_i\rightarrow e_i \in \mathbb{R}^D$ ($i$=1,..$N$) 
to obtain the sequence   $e\in \mathbb{R}^{N \times D}$.
 In this way, spatial and channel features of the $i$-th image patch $x_i$ are transformed into embedded features of the $i$-th embedding vector $e_i$.

\subsection{Transformer Backbone}
We adopt a pyramid transformer backbone Twins \cite{chu2021twins}, which features a scheme of alternated local and global attention. Owning local and global receptive fields captures both short and long range relations. Due to this specific design, Twins obtains great performance across several benchmarks compared with its counterparts. Moreover, it is deployment-friendly, which eases practical application.

Based on the standard transformer module, Twins proposes the spatially separable self-attention
(SSSA) module to reduce the amount of calculation. To achieve so, the input sequences of $l$-th layer $Z^{l-1}$ are reshaped into 2D feature maps first. Each feature map can then be spatially grouped within a local window to compute so-called locally-grouped self-attention (LSA). Successively for Global sub-sampled attention (GSA), Twins adopts subsampling for keys to serve as a representative within each window to cut down the cost. 

Specifically, at the stage of LSA, the feature maps are equally divided into $k_1\times k_2$ sub-windows. And the self-attention computation is locally performed in each sub-window. In this way, the local features can be obtained efficiently. As there are no communications among the sub-windows, the following global attention comes into play. At the stage of GSA, each sub-window generates a single representative through a convolution operation. This representative summarizes the key information of the sub-window. Then a self-attention computation is performed on the representatives of all the sub-windows. In this way, sub-windows can communicate with each other through their representatives, so that global features can be captured. Interleaving the necessary Multi-layer Perceptron (MLP) modules, layer normalization (LN), and residual connections, the structure of the $l$-th layer of the transformer in Twins can be defined as follows:
\begin{align}
	Z'_l &= LSA (LN (Z_{l-1})) + Z_{l-1},\\
	Z''_l &= MLP (LN (Z'_{l})) + Z'_{l},\\
	Z'''_l &= GSA (LN (Z''_{l})) + Z''_{l},\\
	Z_l &= MLP (LN (Z'''_l)) + Z'''_l.
	\label{eq:encoder1}
\end{align}

\subsection{Pyramid Feature Aggregation}


Though the transformer backbone can extract global features, feature maps from high layers still lack detail information that can not be reconstructed by up-sampling. These high-level features are too fuzzy to distinguish the boundaries of different objects, which  makes crowd counting models hard to learn the accurate location information of crowds. To address this problem, we construct a model to make full use of both
 high-level and low-level information.

For the specific stage of the encoding $s$, we  use the output sequence $Z_s\in \mathbb{R}^{N \times d}$ ($s$=1...$T$), which contains global context. Then we reshape the one-dimensional sequence $Z_s$ into a two-dimensional grid of embedded vectors, whose height and width are both $\sqrt{N} $. Regarding the fact that   feature maps from shallow layers contain rich detail information but lack semantics and  those from deep layers are just the opposite. We construct a feature pyramid to aggregate the semantic information from high-level layers with detail information from low-level layers. Specifically, we upsample the feature maps from all  stages to  $\frac{1}{8}$ size of the input image, which is a common choice in most work \cite{li2018csrnet}. This resolution also helps to make fair comparisons with other methods. The structure of Pyramid Feature Aggregation (PFA) is shown in Figure~\ref{fig:pipeline}.

\subsection{Regression Head with Multi-scale Receptive Fields}

\begin{figure}[h]
	\centering
	\subfigure[dilation rate=2, 2, 2]{
		{\includegraphics[width=0.145\linewidth]{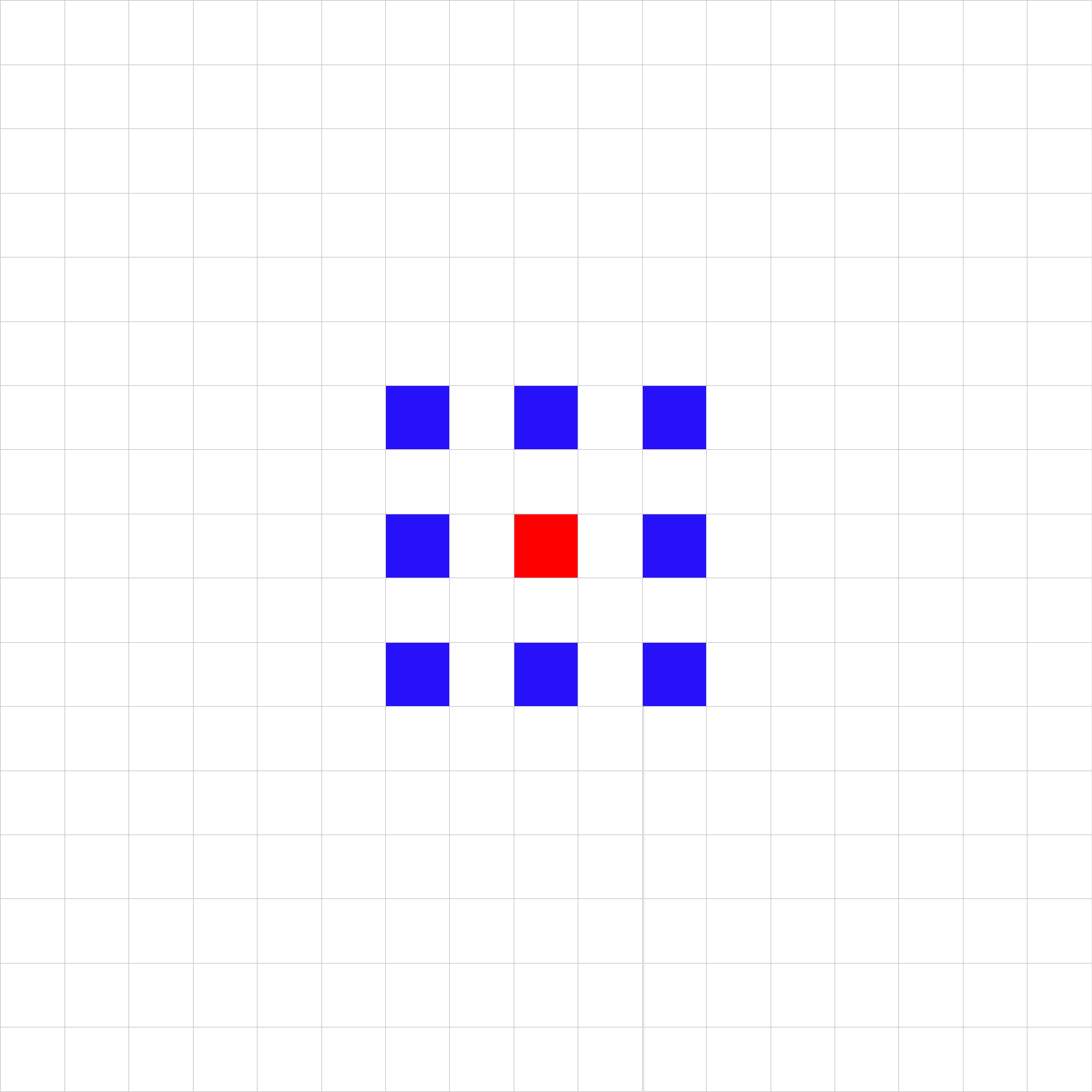}}\,
		{\includegraphics[width=0.145\linewidth]{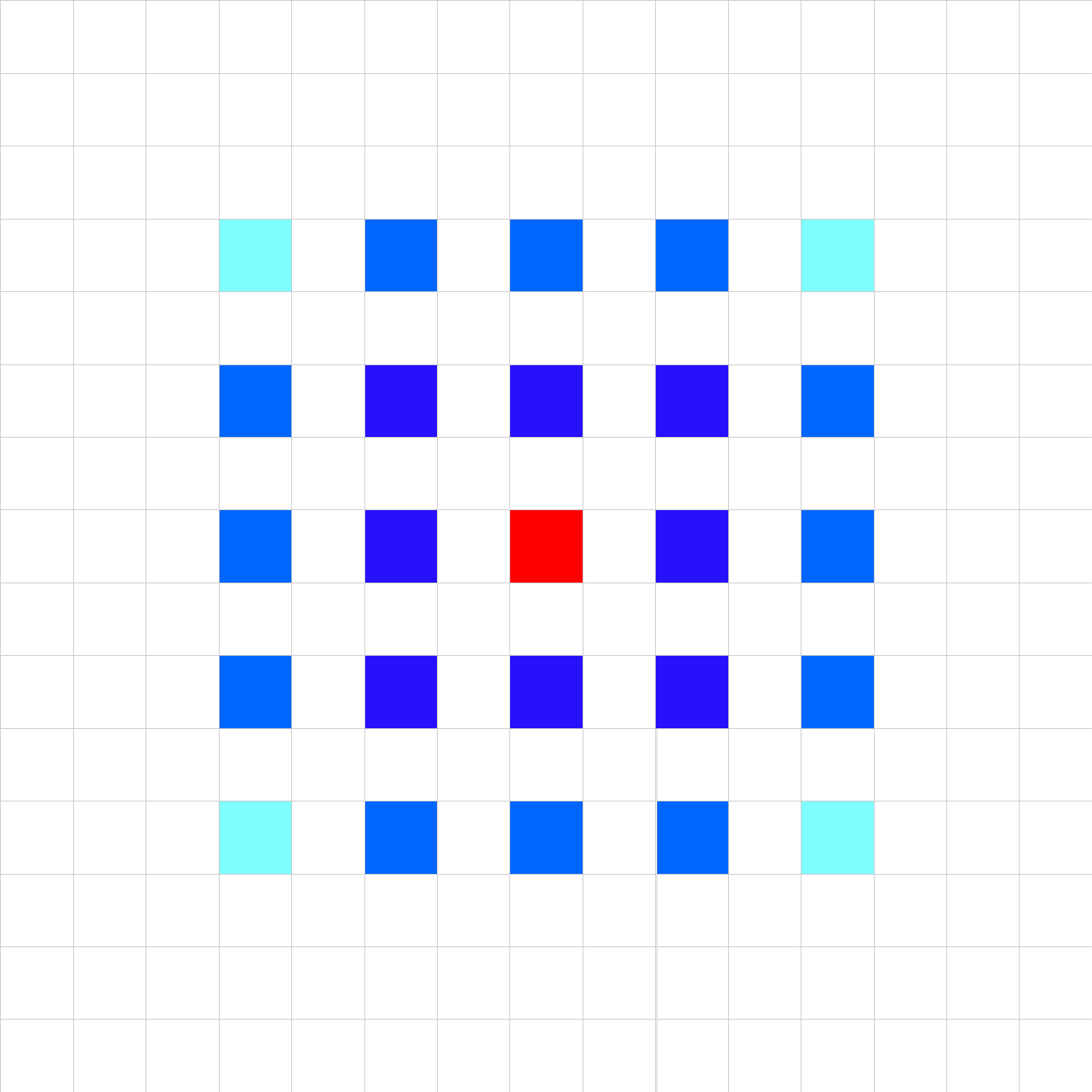}}\,
		{\includegraphics[width=0.145\linewidth]{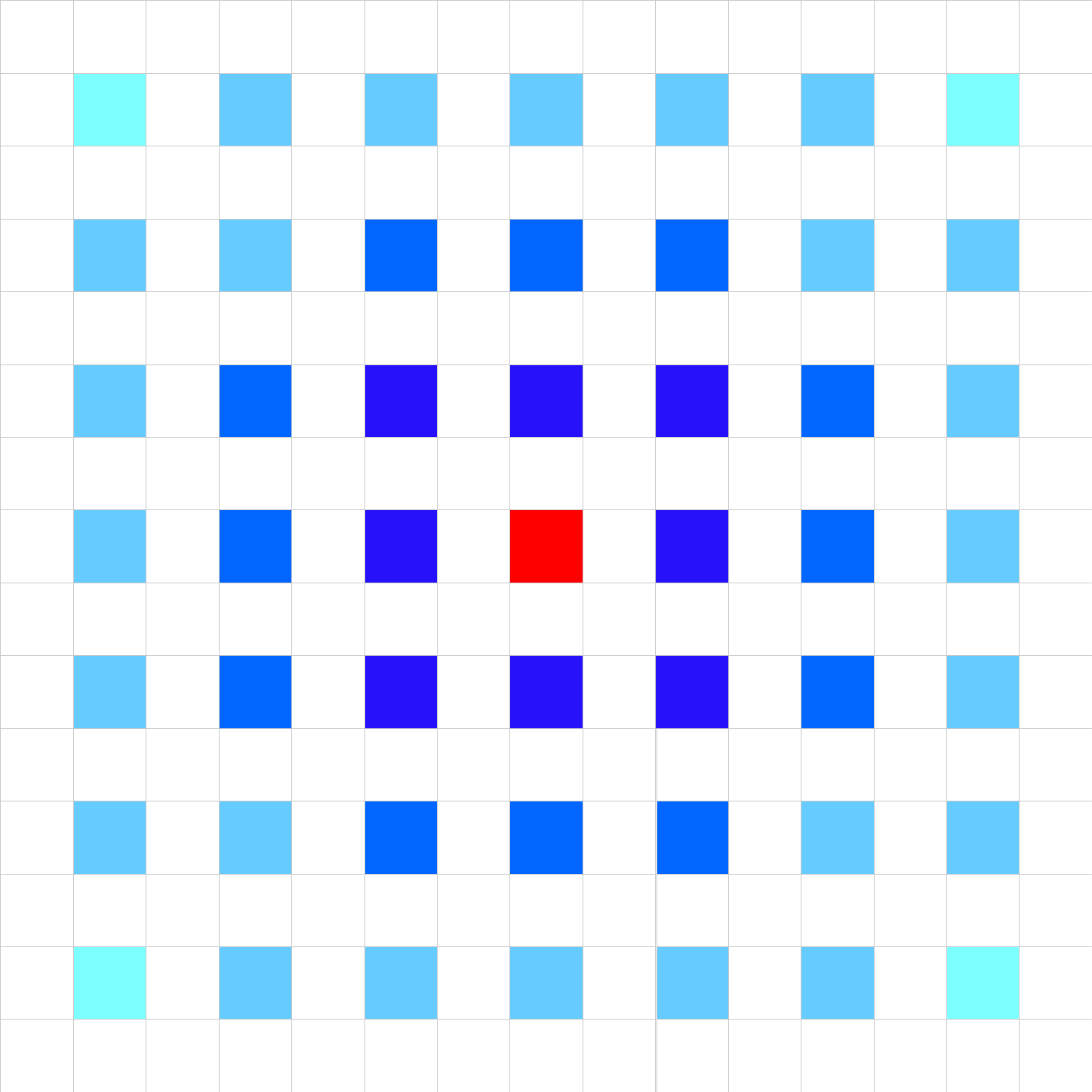}}
	}
	\subfigure[dilation rate=1, 2, 3]{
		{\includegraphics[width=0.145\linewidth]{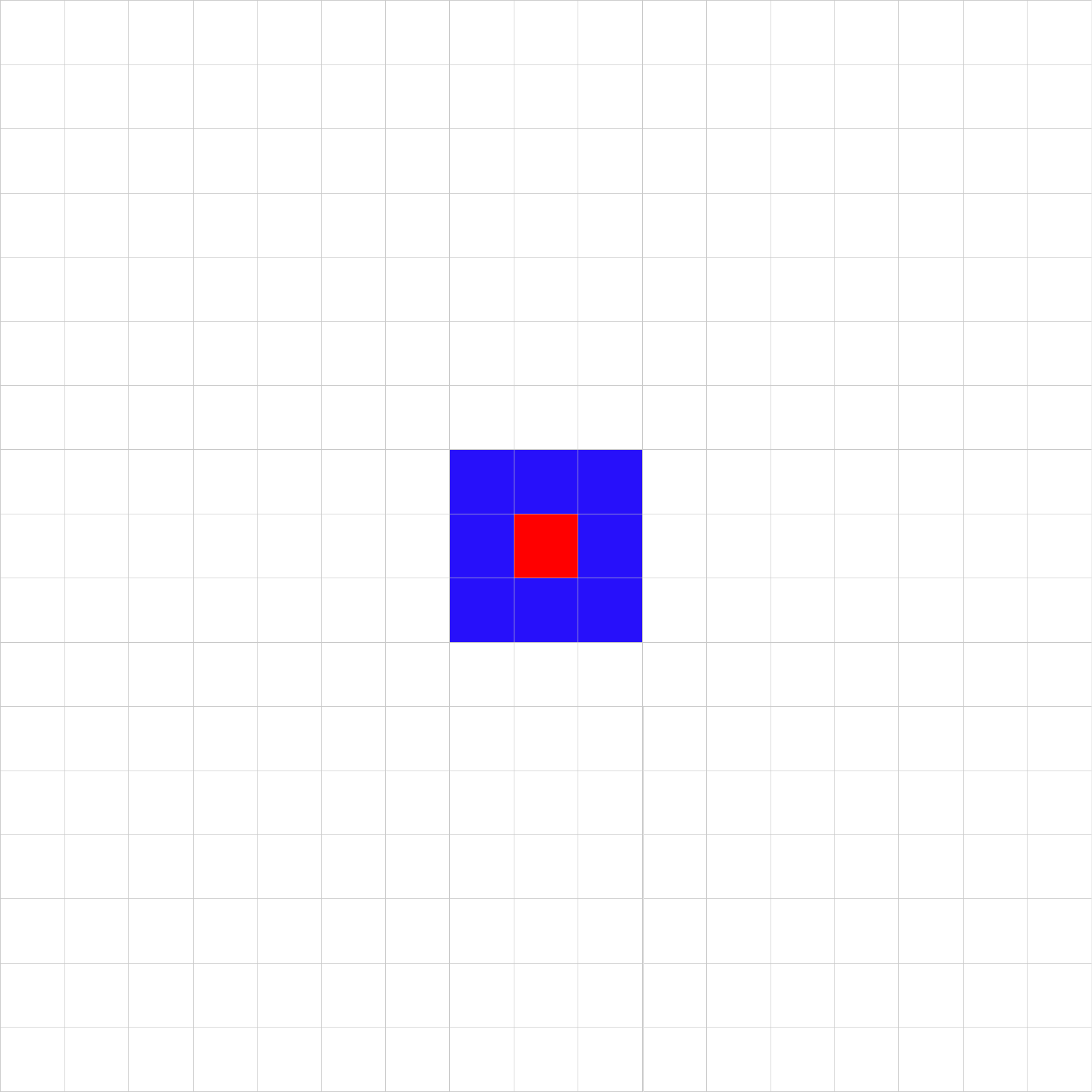}}\,
		{\includegraphics[width=0.145\linewidth]{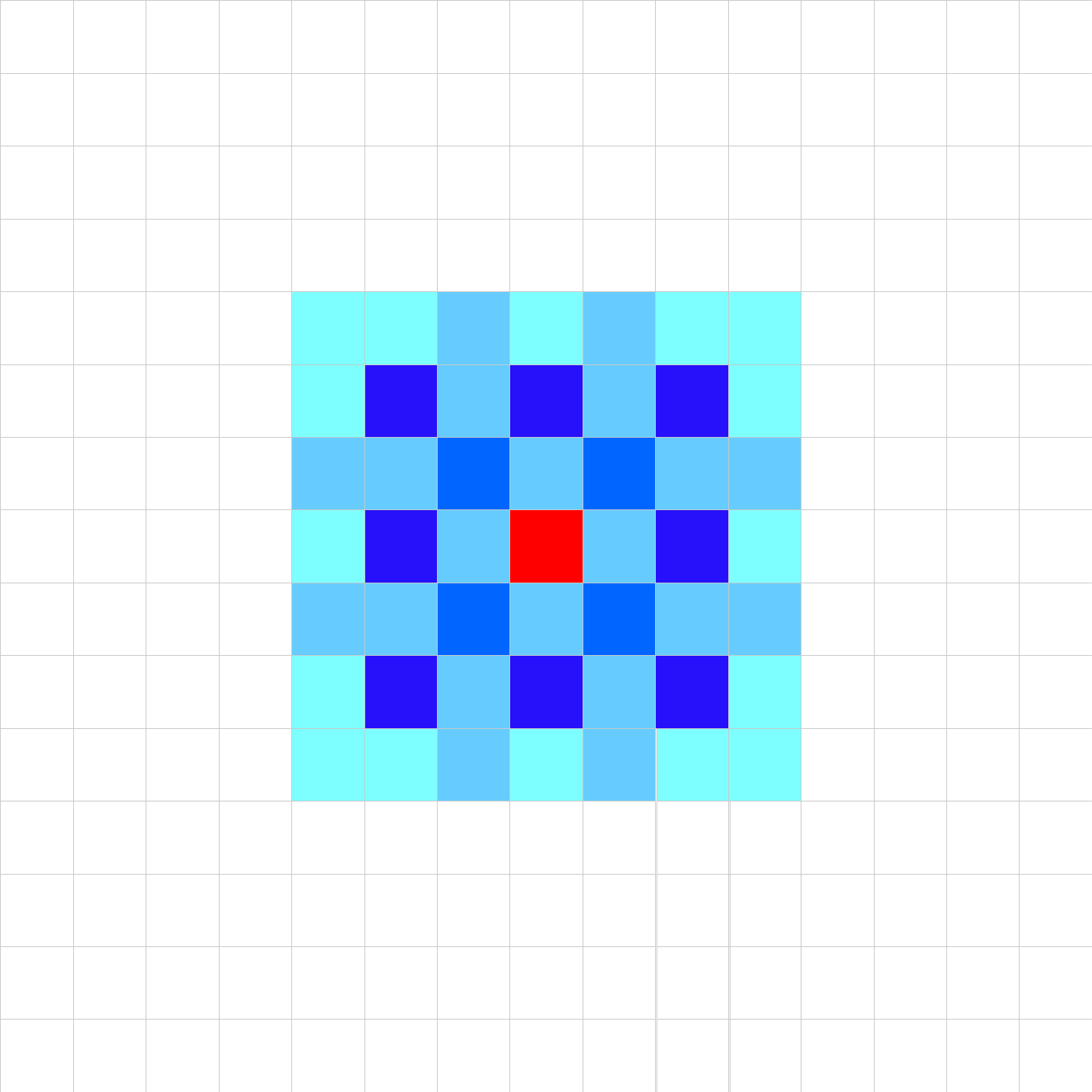}}\,
		{\includegraphics[width=0.145\linewidth]{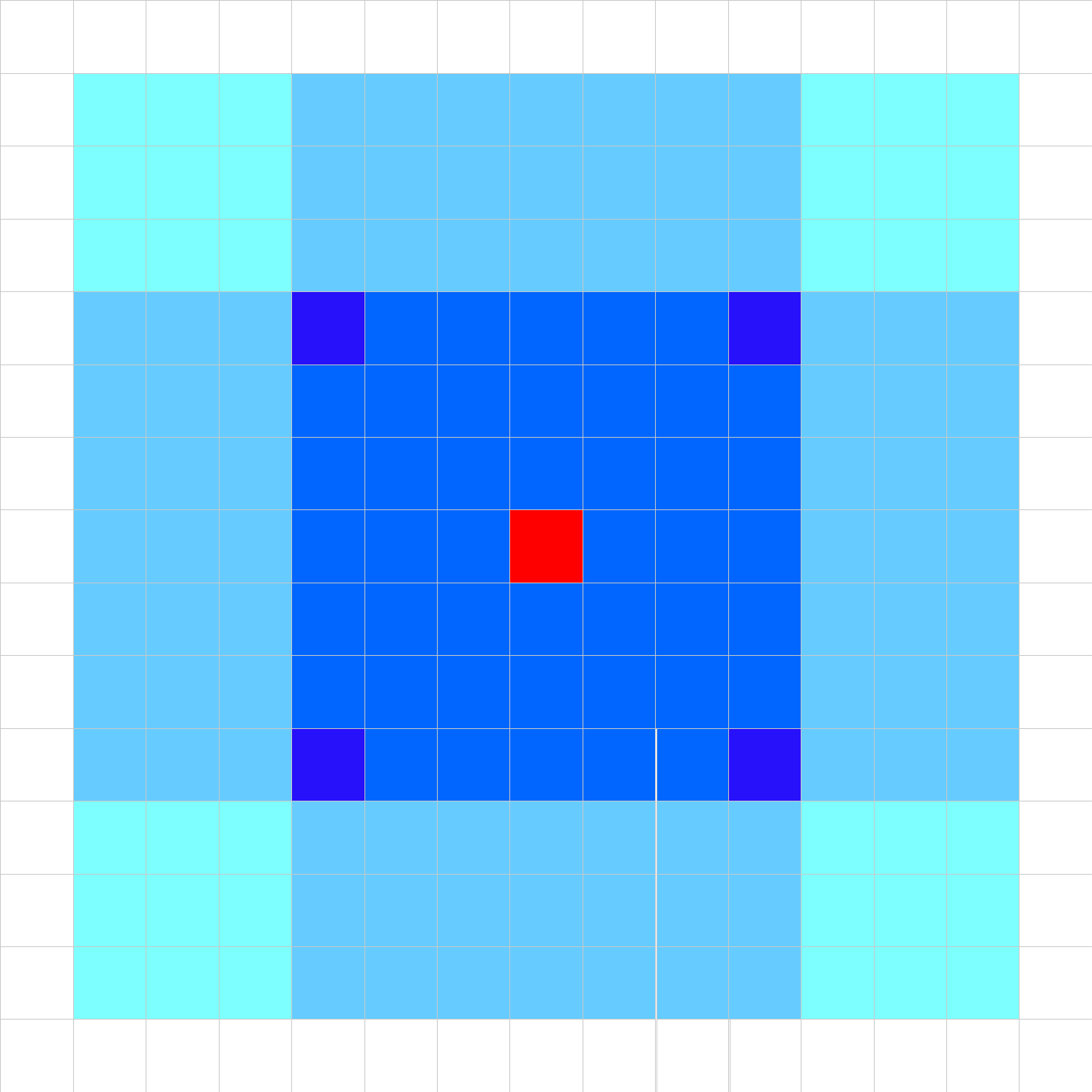}}
	}
	\caption{The gridding effect in stacking DConv layers with the fixed dilation rate. And stacking DConv layers with different dilation rates can avoid this. }
	\label{fig: diated_conv}
\end{figure}

Regarding our transformer-based backbone and PFA block have already captured sufficient global information. Therefore, we only use a simple regression head to regress accurate density maps. Specifically, we construct a module with multi-scale receptive fields to detect the global scale and density variances. A straightforward approach is  stacking dilated convolution (DConv) layers  as \cite{li2018csrnet,yan2019perspective}. DConv layers can enlarge receptive fields while containing the spatial resolution.  However, this design requires careful design of the dilation coefficient at each layer to avoid the 
gridding effect \cite{fang2020face}, where some pixels are  missed in the later convolutions (see Figure~\ref{fig: diated_conv}). The gridding effect has a great impact on the results of crowd counting. Because the scale of crowds beyond the capture is very small, one missing pixel in high-level feature maps will make the regression head ignore several people.   In addition, it is a bit time-consuming to stack layers in depth, not width \cite{szegedy2016inception}. This will increase the training cost and make the model difficult to expand to real-time scenarios.  The experiment result in Table~\ref{tab:MDC_change} shows this design is worse than stacking in width.

  Inspired by DeepLabv3+ \cite{chen2018deeplab} and RFB \cite{liu2018receptive}
   we design a Multi-scale Dilated Convolution (MDC) block by stacking  DConv layers with different dilation rates in parallel.  But different from the ASPP \cite{chen2018deeplab},  MDC is  more lightweight but powerful enough to obtain good performance. 

Specifically, it contains three columns (${C}_{1}$, ${C}_{2}$, ${C}_{3}$) and a shortcut path. And each column  consists of a single convolutional  layer and a dilated convolutional layer. We set the corresponding kernel sizes and dilation rates as small as possible to fit the crowd counting scenes full of small-scale objects. Each convolutional layer is followed by a batch normalization (BN) layer and a ReLU activation function. We  concatenate the output feature maps from each column and add them with a shortcut path to make use of multi-scale features. Finally, we use a ${1 \times 1}$ convolution layer to regress the density map. The structure is shown in Figure~\ref{fig:pipeline}. 

\subsection{Loss Function Design}

We design different loss functions for fully-supervised density regression and weakly-supervised  counting regression. 

\paragraph{Loss function design for the fully-supervised setting.}
Our design is based on a popular loss from \cite{wang2020distribution},
which is formulated by weighted summation of Counting loss, Optimal Transport (OT) loss, and Total Variation (TV) loss. For a predicted density map $D$ and its ground-truth $D'$, the loss function is defined as:
\begin{align}
	\mathcal{L}_{dm}= {L}_{1} (P,G)+\lambda_1 {\mathcal{L}_{OT}}+\lambda_2{\mathcal{L}_{TV}}(D,{D}^{'}),
	\label{eq:dm_loss}%
\end{align}
where $P$ and $G$ denote the crowd number of $D$ and $D'$, respectively. $\lambda_1$ and $\lambda_2$ are the loss coefficient, and they are set to 0.01 and 0.1 in DM-Count \cite{wang2020distribution}.


OT loss benefits the model with a strong fitting ability to minimize the distribution gap between the predicted density map and the ground-truth. However, \citet{wang2020distribution} point out that it can not approximate well the sparse areas of crowds and additionally use an extra TV loss to stabilize.
But TV loss uses the original head annotations of ground-truth, which is not smooth enough to build a robust representation of people. Especially in some sparse scenes, crowds have a larger scale, and it is unreasonable to represent a person by a pixel. 
To address this issue, 
 we utilize the mean square error i.e. $L_2$ to regularize the gap between prediction and smoothed annotation maps. The smooth feature maps are generated by applying the adaptive Gaussian kernels \cite{li2018csrnet}. 
The total loss is written as,
\begin{align}
	\mathcal{L}_{d}= {L}_{1} (P,G)+\lambda_1 {\mathcal{L}_{OT}}+\lambda_2{L}_{2} (D,{D}^{'}).
	\label{eq:density_loss}
\end{align}
In our experiment,  we fix $\lambda_1=0.01$ as \cite{wang2020distribution} and only fine-tune $\lambda_2$\footnote{1.0 is used in our paper.}. Although this may be sub-optimal for our method, it still shows good performance across several benchmark datasets.

\paragraph{Loss function design for the weakly-supervised setting.}
To be robust, we utilize smooth $L_1$ loss instead of $L_1$. Because the number of crowds varies greatly in different images, and $L_1$ is sensitive to outliers. Our weakly-supervised loss function is defined as:
\begin{align}
	\mathcal{L}_{c}={{smooth}_{{L}_{1}}(D,{D}^{'})}. 
	\label{eq:counting_loss}
\end{align}
Experimental results in Table~\ref{tab:Loss_ablation} also show this design can bring boosted performance.

\section{Experiments }

\subsection{Implementation Details.}
\paragraph{Dataset.}
We evaluate our method across five benchmarks, including UCF\_CC\_50~\cite{idrees2013multi}, ShanghaiTech Part A and Part B~\cite{zhang2016single}, UCF\_QNRF~\cite{idrees2018composition}, and NWPU-Crowd~\cite{wang2020nwpu}. These datasets differ in image resolution ratios, quantities, crowding degree, and color spaces. 
\paragraph{Training setting and hyper-parameter.}
To make fair comparisons, the transformer-based backbone is the official Twins-SVT-large model, which is pretrained on the ImageNet 1k dataset \cite{deng2009imagenet}. We only use random cropping and random horizontal flipping as data augmentations  for all experiments, which strictly follows \cite{wang2020distribution,ma2019bayesian}. The crop size of both ST\_Part B and UCF\_QNRF  is 512 and is changed to 256 for UCF\_CC\_50 and ST\_Part A. We use AdamW \cite{loshchilov2018decoupled} with a batch size of 8 because it is more suitable for training transformer-based models. The initial learning rate is set to 1e-5. We use  l2 regularization of 0.0001 to avoid over-fitting.  All experiments are conducted using PyTorch on a single 32G Tesla V100 GPU.

\paragraph{Evaluation metric.}
Previous works in crowd density estimation use Mean Absolute Error (MAE) and Root Mean Squared Error (RMSE or MSE for short) as evaluation metrics in CSRNet. Besides, mean Normalized Absolute Error (NAE) is an extra metric from \cite{wang2020nwpu}. They can be formulated as follows:

\begin{equation}
	MAE=\frac{1}{N}\sum_{i=1}^{N}\left |P_{i}-G_{i} \right|,
	\label{eq:mae}
\end{equation}
\begin{equation}
	MSE=\sqrt{\frac{1}{N}\sum_{i=1}^{N}\left |P_{i}-G_{i} \right|^{2}},
	\label{eq:mse}
\end{equation}
\begin{equation}
NAE = \frac{1}{N}\sum\limits_{i = 1}^N {\frac{\left| {{P_i} - {{G}_i}} \right|}{{G_i}}},
\label{eq:nae}
\end{equation}
where $N$ is the number of testing images, $P_{i}$ and $G_{i}$ are the predicted and ground-truth number of crowds in the $i$-th image, respectively.  
\subsection{Comparisons with State-of-the-art Methods}

\paragraph{Performace on UCF$\_$CC$\_$50.} This dataset shows a lot of challenges. It randomly collects only 50 gray images with  serious perspective distortions from the Internet.  We report the result in Table~\ref{tab:qab_maemse}. Our method surpasses ASNet by 3.5\% in MAE and  CAN by 3.8\% in MSE. It indicates the transformer based on self-attention mechanism is robust to visual distortion and the deficiency in RGB information. 
\paragraph{Performace on ShanghaiTech.} The result is shown in Table~\ref{tab:qab_maemse}. For part A, the images are randomly crawled from the Internet. The number of people in these images varies largely with a wide range. 
 For part B, the images are captured by the surveillance cameras in the streets of Shanghai. These images have dramatic intra-scene scale and density variances. CNN-based methods lack context modeling ability to deal with them. Our method obtains new state-of-the-art, which outperforms the previous best method P2PNet \cite{song2021rethinking} without complicated designs.
\paragraph{Performace on UCF$\_$QNRF.}The images of this dataset are collected from different websites. These images contain large diversity both in scenes and image sizes. Most of the objects in the pictures are small in scale.  The result is shown in Table~\ref{tab:qab_maemse}. Our method outperforms other methods again. That is because our PFA can contain more detail information, which is helpful for our model to detect small objects. And our MDC can better capture multi-scale features and global context information from the transformer to regress the crowd number.
\begin{table*}[tb!]
	\centering
	\setlength{\tabcolsep}{1.1mm}
	\begin{tabular}{ |l|c|cc|cc|cc|cc|cc| }
		\hline
		{\multirow{2}{*}{Method}} &{\multirow{2}{*}{Venue}}&\multicolumn{2}{c|}{Label}&\multicolumn{2}{c|}{UCF\_CC\_50} & \multicolumn{2}{c|}{ST\_Part A} & \multicolumn{2}{c|}{ST\_Part B} & \multicolumn{2}{c|}{UCF\_QNRF}\\
		\cline{3-12}
		&&Location&Number& MAE & MSE & MAE & MSE&MAE&MSE&MAE&MSE \\
		\hline
		CAN~\shortcite{liu2019context}&CVPR19&$\surd$&$\surd$&212.2&243.7&62.3&100.0&7.8&12.2& 107.0 & 183.0\\
		SFCN~\shortcite{wang2019learning}&CVPR19&$\surd$&$\surd$&214.2&318.2&59.7&95.7&7.4&11.8& 102.0 & 171.0\\
		PACNN~\shortcite{shi2019revisiting}&CVPR19&$\surd$&$\surd$&241.7&320.7&62.4&102.0&7.6&11.8& - & -\\
		S-DCNet~\shortcite{xiong2019open} &ICCV19&$\surd$&$\surd$ &204.2&301.3&58.3 & 95.0 & 6.7& 10.7& 104.4 & 176.1 \\
		DSSI-Net~\shortcite{liu2019crowd}&ICCV19 &$\surd$&$\surd$&216.9&302.4&60.6&96.0 &6.8&10.3& 99.1 & 159.2\\
		BL~\shortcite{ma2019bayesian} &ICCV19 &$\surd$&$\surd$&229.3&308.2& 62.8& 101.8& 7.7 & 12.7& 88.7 & 154.8\\
		RPNet~\shortcite{yang2020reverse} &CVPR20&$\surd$&$\surd$&-&-&61.2&96.9&8.1&11.6& - & -\\
		ASNet~\shortcite{jiang2020attention}&CVPR20&$\surd$&$\surd$&174.8&251.6&57.8&90.1&-&-& 91.5 & 159.7\\
		LibraNet~\shortcite{liu2020weighing} &ECCV20&$\surd$&$\surd$&181.2&262.2&55.9&97.1&7.3&11.3& 88.1 & 143.7\\
		AMRNet~\shortcite{liu2020adaptive} &ECCV20&$\surd$&$\surd$&184.0&265.8&61.5&98.3&7.0&11.0& 86.6 & 152.2\\
		NoisyCC~\shortcite{wan2020modeling}&NeurIPS20&$\surd$&$\surd$&-&-&61.9&99.6&7.4&11.3& 85.8 & 150.6\\
		DM-Count~\shortcite{wan2020modeling}&NeurIPS20&$\surd$&$\surd$&211.0&291.5&59.7&95.7&7.4&11.8& 85.6 & 148.3\\
		GL\shortcite{Wan_2021_CVPR}&CVPR21 &$\surd$&$\surd$& -&-&61.3&95.4& 7.3&11.7& 84.3&147.5 \\
		SUA-Fully \shortcite{meng2021spatial}&ICCV21 &$\surd$&$\surd$& -&-&66.9&125.6& 12.3&17.9& 119.2&213.3 \\
		P2PNet \shortcite{song2021rethinking} &
		 ICCV21&$\surd$&$\surd$&172.7&256.2&52.7&85.1&6.3&\textbf{9.9}&85.3&154.5\\
		BCCT \shortcite{sun2021boosting} &
		arxiv21&$\surd$&$\surd$&-&-&53.1&\textbf{82.2}&7.3&11.3&83.8&143.4\\	
		\textbf{CCTrans (ours)}~&-&$\surd$&$\surd$&\textbf{168.7}&\textbf{234.5}&\textbf{52.3}&84.9&\textbf{6.2}&\textbf{9.9}&\textbf{82.8}&\textbf{142.3}\\
		\hline
		Yang et al.~\shortcite{yang2020weakly}*&ECCV20&\xmark&$\surd$ &-&-& 104.6 & 145.2 & 12.3 &21.2& - & -\\
		MATT~\shortcite{lei2021towards}*&PR21&\xmark&$\surd$ &355.0&550.2& 80.1 & 129.4 & 11.7 &17.5& - & -\\
		TransCrowd~\shortcite{liang2021transcrowd}*&arxiv21&\xmark&$\surd$&-&-&66.1&105.1&9.3&16.1 & 97.2 & 168.5\\
		\textbf{CCTrans (ours)*}~&-&\xmark&$\surd$&\textbf{245.0}&\textbf{343.6}&\textbf{64.4}&\textbf{95.4}&\textbf{7.0}&\textbf{11.5}&\textbf{92.1}&\textbf{158.9}\\
		\hline
	\end{tabular}
	\caption{Comparison with state-of-the-art methods on UCF\_CC\_50, ShanghaiTech Part A and B, and
		UCF-QNRF datasets. * represents the weakly-supervised method. Our method achieves new state-of-the-art results on several standard benchmarks. }
	\label{tab:qab_maemse}
\end{table*}
\paragraph{Performance on NWPU-Crowd.} 
  Different from ShanghaiTech Part B which has a dramatic intra-scene scale and density variance, this dataset has a serious inter-scene scale and density variance. On this new and challenging dataset, our method surpasses all previous methods with a large margin. The result is shown in Table~\ref{tab:NWPU_crowd}. Our method obtains 38.6 MAE on the validation dataset, which is  14.4 lower than the concurrent work BCCT \cite{sun2021boosting}. We also evaluate our method on the test set by uploading the results to the official server and our method obtains 69.3 MAE, which is 8.1 lower than the previous best method P2PNet \cite{song2021rethinking}. Note that we  rank No.1 on the leader-board without bells and whistles.
\begin{table}[ht]
	\centering
	\smallskip
	\setlength{\tabcolsep}{0.8mm}
	\resizebox{0.475\textwidth}{!}{
		\begin{tabular}{ |l|cc| cc| ccc| }
			\hline
			{\multirow{2}{*}{Method}} &\multicolumn{2}{c|}{Label} & \multicolumn{2}{c|}{Val} & \multicolumn{3}{c|}{Test} \\
			\cline{2-8}
			&L&N& MAE & MSE&MAE&MSE&NAE \\
			\hline
			MCNN\shortcite{zhang2016single}&$\surd$&$\surd$&218.5&700.6&232.5&714.6&1.063\\
			CSRNet \shortcite{li2018csrnet}&$\surd$&$\surd$&104.9&433.5 & 121.3 & 387.8&0.604\\
			CAN \shortcite{liu2019context}&$\surd$&$\surd$&93.5&489.9 & 106.3 & 386.5&0.295\\ 
			BL \shortcite{ma2019bayesian}&$\surd$&$\surd$&93.6&470.4 & 105.4 & 454.2&0.203\\
			DM-Count \shortcite{wan2020modeling} &$\surd$&$\surd$&70.5&357.6& 88.4& 388.6&0.169 \\
			
			GL \shortcite{Wan_2021_CVPR} &$\surd$&$\surd$ &-&-& 79.3& 346.1&0.180 \\
			P2PNet \shortcite{song2021rethinking}&$\surd$&$\surd$ &-&-& 77.4& 362.0&- \\
			BCCT \shortcite{sun2021boosting}&$\surd$&$\surd$&53.0&170.3& 82.0& 366.9&0.164 \\
			\textbf{CCTrans (ours)} &$\surd$&$\surd$&\textbf{38.6} &\textbf{87.8} & \textbf{69.3} & \textbf{299.4}&\textbf{0.135}\\
			
			\hline
			TransCrowd~\shortcite{liang2021transcrowd}* &\xmark&$\surd$&88.4 &400.5 & 117.7 & 451.0&0.244\\
			\textbf{CCTrans (ours)}* &\xmark&$\surd$&\textbf{48.6} &\textbf{121.1} & \textbf{79.8} & \textbf{344.4}&\textbf{0.157}\\
			\hline
		\end{tabular}
	}
	\caption{Comparisons on the validation and testing set of NWPU-Crowd. Our method achieves new state-of-the-art results on this large benchmark with clear margins.}
	\label{tab:NWPU_crowd}
\end{table}
\paragraph{Comparison with other Transformer-based approaches.} There are two concurrent transformer-based works. One is TransCrowd \cite{liang2021transcrowd}, which is designed for solving crowd counting under the weakly-supervised setting. It utilizes ViT \cite{dosovitskiy2021an} to extract limited features from only one stage. And it uses an additional regression token and global average pooling to perform counting regression, respectively. This design can not make full use of the advantage of global attention. Table~\ref{tab:qab_maemse} and ~\ref{tab:NWPU_crowd} show that our method outperforms it with clear margins across several benchmarks. Another is BCCT \cite{sun2021boosting}, which also utilizes ViT as the backbone and designs various attention mechanisms and complicated auxiliary losses to boost the performance. Nevertheless, our method  outperforms it with a clear advantage on most benchmarks, which is shown in Table~\ref{tab:qab_maemse} and ~\ref{tab:NWPU_crowd}.
\begin{figure}[ht]
	\centering
	\includegraphics[width=0.48\textwidth]{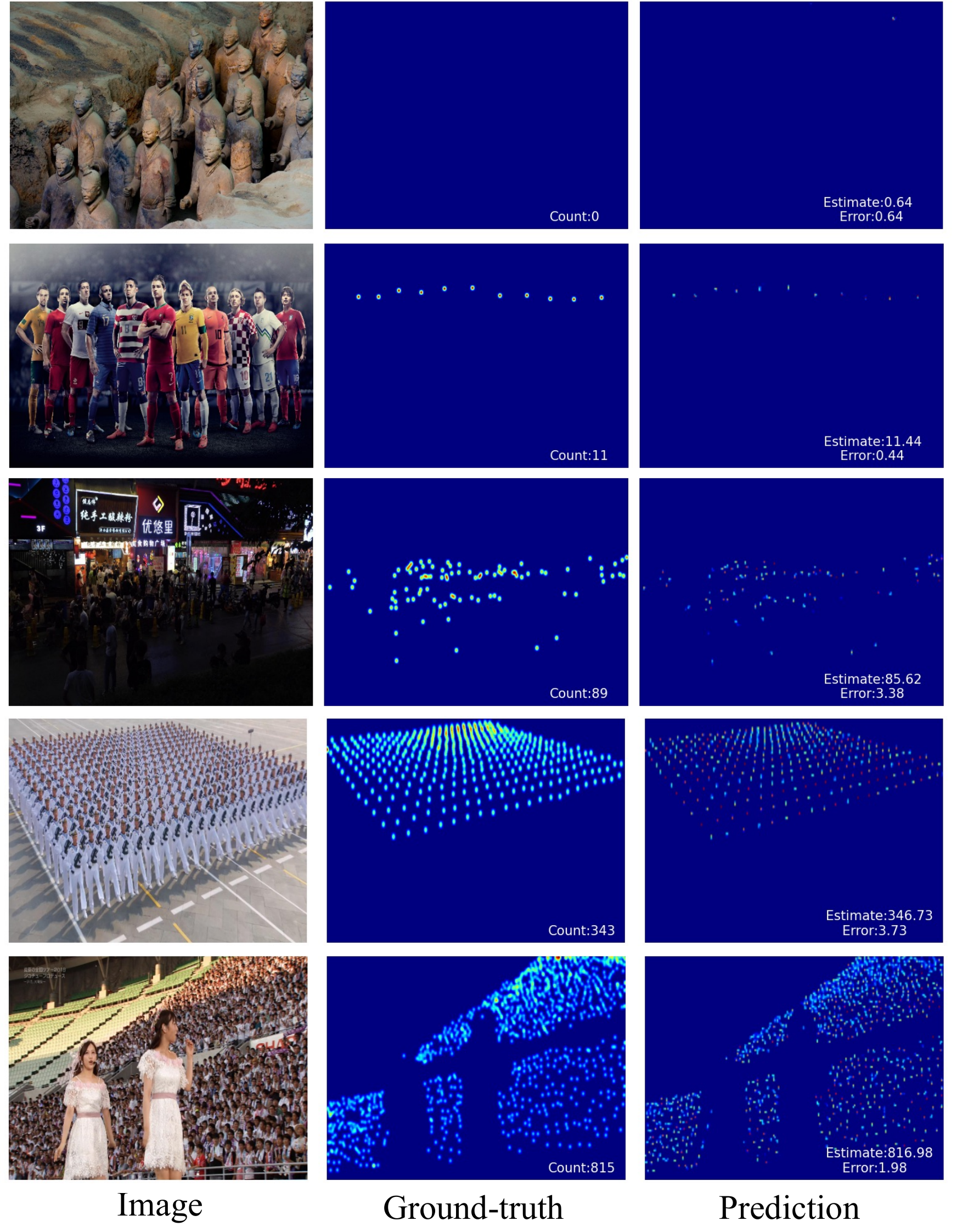}
	\caption{Visualization results of CCTrans on the most-challenged NWPR-Crowd. Five images in different scenes with varied scale, density, and illumination differences show the fitting ability of CCTrans.}
	\label{fig:Viz}
\end{figure}
\subsection{Visualization}
We show some visualization results in  Figure~\ref{fig:Viz}. More analysis and results are in Figure~\ref{fig:moreViz} of the supplementary material.
\subsection{Ablation Study}
We make extensive ablation experiments on ST$\_$Part A and ST$\_$Part B to evaluate the contribution of each component. For simplicity, we use L and N to denote location and number label, respectively.

\paragraph{PFA.} We evaluate the performance of PFA using features from different stages in Table~\ref{tab:PFA_ablation}. The deeper layer learns high-level semantic features which are critical to the crowd counting task. $F_1$, $F_2$, and $F_3$ are the output features from three stages in the transformer-based backbone, and they are named from shallow layers to deep ones. It's interesting that using the feature from the last stage already outperforms most of the state-of-the-art methods. Aggregating features from other stages can provide rich information which is lost in the last stage, thus achieving better performance. 
\begin{table}[tb!]
	\centering
	\smallskip
	\setlength{\tabcolsep}{1.0mm}
	\resizebox{0.45\textwidth}{!}{
		\begin{tabular}{ |l|cc|cc|cc| }
			\hline
			{\multirow{2}{*}{Features in PFA}} &\multicolumn{2}{c|}{Label} & \multicolumn{2}{c|}{ST\_Part A} & \multicolumn{2}{c|}{ST\_Part B} \\
			\cline{2-7}
			&L&N & MAE & MSE&MAE&MSE \\
			\hline
			
			$F_1$&$\surd$&$\surd$&81.5&126.9&9.3&15.0 \\
			$F_2$&$\surd$&$\surd$&56.2&93.4&6.7&10.4 \\
			$F_3$&$\surd$&$\surd$&54.8&86.2&6.8&10.7 \\
			\textbf{$F_1$+$F_2$+$F_3$(CCTrans)}&$\surd$&$\surd$&\textbf{52.3}&\textbf{84.9}&\textbf{6.2}&\textbf{9.9} \\
			\hline
		\end{tabular}
	}
	\caption{Feature aggregation from three stages achieves the best performance.}
	\label{tab:PFA_ablation}
\end{table}
\paragraph{MDC.} In Table~\ref{tab:MDC_ablation}, we evaluate the performance of MDC using different columns with a shortcut path. Our shortcut path only uses a $1\times1$ Conv layer to adapt the input feature maps of the following layer. It is included in all the experiments by default. We use $C_1$, $C_2$, and $C_3$ to denote the three columns from up to bottom of the regression head in Figure~\ref{fig:pipeline}. It's interesting to see that using  $C_1$ or $identity$ only already outperforms most of the recent SOTAs. Proper aggregation of all columns can further improve the performance, which validates the necessity of MDC.
\begin{table}[tb!]
	\centering
	\smallskip
	\setlength{\tabcolsep}{1.0mm}
	\resizebox{0.45\textwidth}{!}{
		\begin{tabular}{ |l|cc|cc|cc| }
			\hline
			{\multirow{2}{*}{Column Number}} &\multicolumn{2}{c|}{Label} & \multicolumn{2}{c|}{ST\_Part A} & \multicolumn{2}{c|}{ST\_Part B} \\
			\cline{2-7}
			&L&N & MAE & MSE&MAE&MSE \\
			\hline
			
			$identity$&$\surd$&$\surd$&55.8&89.8&6.9&11.3 \\
			$C_1$&$\surd$&$\surd$&55.9&93.3&6.8&11.5 \\
			$C_2$&$\surd$&$\surd$&57.9&95.2&7.5&12.3 \\
			$C_3$&$\surd$&$\surd$&56.5&91.3&7.3&11.8 \\
			\textbf{$C_1$+$C_2$+$C_3$(CCTrans)}&$\surd$&$\surd$&\textbf{52.3}&\textbf{84.9}&\textbf{6.2}&\textbf{9.9} \\
			\hline
		\end{tabular}
	}
	\caption{Multi-scale columns achieve the best performance.}
	\label{tab:MDC_ablation}
\end{table}
\begin{table}[tb!]
	\centering
	\smallskip
	\setlength{\tabcolsep}{1.0mm}
	\resizebox{0.475\textwidth}{!}{
		\begin{tabular}{ |l|cc|cc|cc|c|}
			\hline
			{\multirow{2}{*}{Strategy}} &\multicolumn{2}{c|}{Label} & \multicolumn{2}{c|}{ST\_Part A} & \multicolumn{2}{c|}{ST\_Part B} &{\multirow{2}{*}{Training Time(s)}}\\
			\cline{2-7}
			&L&N & MAE & MSE&MAE&MSE& \\
			\hline
			$identity$&$\surd$&$\surd$&55.8&89.8&6.9&11.3 &\textbf{18.6}\\
			$C_1'$\_$C_2'$\_$C_3'$&$\surd$&$\surd$&55.5&94.6&7.3&11.7 &24.9\\
			$C_1$\_$C_2$\_$C_3$&$\surd$&$\surd$&54.4&89.7&6.8&10.8 &22.1\\
			\textbf{$C_1$+$C_2$+$C_3$(CCTrans)}&$\surd$&$\surd$&\textbf{52.3}&\textbf{84.9}&\textbf{6.2}&\textbf{9.9} &19.3\\
			\hline
		\end{tabular}
	}
	\caption{Stacking DConv layers in width achieves the best performance without much time-consuming.}
	\label{tab:MDC_change}
\end{table}

To validate the rationality of our design for MDC, we also try the different stacking strategies in Table~\ref{tab:MDC_change}. Here $identity$ and $C_1$+$C_2$+$C_3$ have the same meaning as in Table~\ref{tab:MDC_ablation}. We use $C_1$\_$C_2$\_$C_3$ to denote stacking the three columns in depth and $C_1'$\_$C_2'$\_$C_3'$  to describe the dilation rates of the corresponding column being set to 2. Note that the shortcut path is used by default. We can observe that stacking DConv layers with the same dilation rate is bad for performance. Both indicators have decreased to varying degrees, compared with the $identity$. This is possibly due to the above-mentioned gridding effect caused by this stacking strategy, which has an adverse impact on the feature extraction of small-scale objects. Stacking DConv layers with varied dilation rates can effectively solve this problem to improve the results. But stacking too deep Conv layers to strengthen the local features seems not helpful for crowd counting. Besides, stacking these blocks in depth also increases the training time by 15\% to 20\%. It is also worth noticing that MDC is lightweight which brings only less than 5\% additional time cost.
\paragraph{Loss functions.}
We also quantitatively evaluate the improvement from our loss design and report the result in  Table~\ref{tab:Loss_ablation}. As for the fully-supervised setting, it brings in 2.5 and 2.1 lower MAE for Part A and B respectively. Compared with the $L_1$ loss, the design of smooth $L_1$  decreases 0.7 and 1.3 MAE for the ShanghaiTech A and B dataset under the weakly-supervised setting.

\begin{table}[tb!]
	\centering
	\smallskip
	\setlength{\tabcolsep}{1.0mm}
	\resizebox{0.45\textwidth}{!}{
		\begin{tabular}{ |l|cc|cc|cc| }
			\hline
			{\multirow{2}{*}{Loss}} &\multicolumn{2}{c|}{Label} & \multicolumn{2}{c|}{ST\_Part A} & \multicolumn{2}{c|}{ST\_Part B} \\
			\cline{2-7}
			&L&N & MAE & MSE&MAE&MSE \\
			\hline

			$\mathcal{L}_{c}$+$\mathcal{L}_{OT}$+$\mathcal{L}_{Tv}$&$\surd$&$\surd$&54.8&86.6&8.3&12.6 \\
			\textbf{$\mathcal{L}_{d}'$(CCTrans,$\lambda_2$=0.1)}&$\surd$&$\surd$&53.7&\textbf{81.2}&6.3&10.1\\
			\textbf{$\mathcal{L}_{d}$(CCTrans,$\lambda_2$=1)}&$\surd$&$\surd$&\textbf{52.3}&84.9&\textbf{6.2}&\textbf{9.9}\\
			\hline
			$\mathcal{L}_{1}$&\xmark&$\surd$&65.1&102.7&8.3&13.7 \\
			\textbf{$\mathcal{L}_{c}$(CCTrans*)}&\xmark&$\surd$&\textbf{64.4}&\textbf{94.4}&\textbf{7.0}&\textbf{11.5} \\
			\hline
		\end{tabular}
	}
	\caption{Our loss functions achieve better performance.}
	\label{tab:Loss_ablation}
\end{table}
\paragraph{Sensitivity about $\lambda_2$.} We investigate the influence of $\lambda_2$ and report the result in Table~\ref{tab:Loss_ablation}. We use $\lambda_2=1$ as our default setting because it outperforms better than $\lambda_2=0.1$ with weak advantages. It indicates that the performance of our method is somewhat robust to the choice of $\lambda_2$.
\paragraph{Component analysis of CCTrans.}  Table~\ref{tab:pipeline_ablation} shows the component analysis of our pipeline. As for the ShanghaiTech  dataset, both PFA and MDC significantly contribute to the final  MAE metric.  $L_d$ plays an important role in decreasing the MSE metric.  
\begin{table}[tb!]
	\centering
	\smallskip
	\setlength{\tabcolsep}{1.1mm}
	\resizebox{0.47\textwidth}{!}{
		\begin{tabular}{ |l|cc|cc|cc| }
			\hline
			{\multirow{2}{*}{Method}} &\multicolumn{2}{c|}{Label} & \multicolumn{2}{c|}{ST\_Part A} & \multicolumn{2}{c|}{ST\_Part B} \\
			\cline{2-7}
			&L&CN & MAE & MSE&MAE&MSE \\
			\hline
			
			Baseline+$L_{dm}$&$\surd$&$\surd$&57.9&94.3&7.7&12.8 \\
			Baseline+$L_{d}$&$\surd$&$\surd$&57.4&89.6&7.6&11.6 \\
			Baseline+$L_{d}$+PFA&$\surd$&$\surd$&54.8&86.2&6.9&11.3 \\
			Baseline+$L_{d}$+PFA+MDC&$\surd$&$\surd$&\textbf{52.3}&\textbf{84.9}&\textbf{6.2}&\textbf{9.9} \\
			\hline
		\end{tabular}
	}
	\caption{Pipeline component analysis of CCTrans.}
	\label{tab:pipeline_ablation}
\end{table}

\section{Related work}
\subsection{CNN-based methods}
Before the popularity of deep learning, researchers use traditional machine learning methods to solve crowd counting \cite{2012Pedestrian,viola2001robust}. But these methods are only adapted to sparse crowd scenes because these detectors have poor performance to distinguish the occluded bodies. Then with the development of deep learning, CNN-based designs are the de facto solutions for crowd counting. These methods can be broadly classified into two categories, i.e. detection-based and density-based methods.
\paragraph{Detection-based methods.}Point-in-box-out \cite{liu2019point} utilizes CNN to construct a detection model to predict bounding boxes for each person, and the number of the boxes represents the number of people. But it can not deal with the occlusion that existed in crowds. RGBD-Net \cite{lian2019density} also constructs a detection model but uses the additional depth information to detect the people occluded by others. However, it not only needs additional data but also gets limited performance in congested scenes.
\paragraph{Density-based methods.}To achieve high performance in congested scenes, density-based approaches are proposed to regress a density map from the input image \cite{pham2015count}. The density map reflects the probability estimation of crowds. The value of each pixel represents the probability of people at that location. The sum of pixel values represents the number of people. Since CNN has a strong ability to extract local features for accurate prediction in local regions, the results are improved a lot. The existing literature focuses on more challenging problems (e.g. scale variance problem and density problem). For the scale variance problem, MCNN \cite{zhang2016single} and Switch-CNN \cite{sam2017switching} design multi-column architectures to extract features in different scales. CFF also constructs different branches for different tasks to refine the model. CAN \cite{liu2019context} and PGCNet \cite{yan2019perspective} propose scale attention mechanisms, which use different sizes of Gaussian kernels or convolution kernels to generate density maps for scale-varied regions. ADCrowdNet uses an auxiliary network to generates an attention map for the crowds in images. Besides, BL \cite{ma2019bayesian} designs a novel loss function, which calculates the specific loss based on the scale of crowds. For the density problem, ASNet \cite{jiang2020attention} divides the crowds into different density levels and assigns the corresponding predictions with different weights based on the density level. RAZNet \cite{liu2019recurrent} introduces additional localization branches to learn position information of crowds, which reduces the estimation errors in varied density regions. RPNet \cite{yang2020reverse} uses transformations between the elliptic coordinates and cartesian coordinates to alleviate the density variance and scale variance simultaneously. For the weakly-supervised method, MATT \cite{lei2021towards} only uses a small amount of point-level annotations to make crowd counting. Yang et al.\cite{yang2020weakly} directly regress the crowd numbers of input images without point-level annotations data.

\subsection{Vision Transformer}


A standard transformer encoder of each stage contains $L$ layers of Multi-head self-attention (MSA) and Multi-layer Perceptron (MLP) blocks. And each layer $l$ has layer normalization (LN) and residual connections. 
For the specific output of the $ (l-1)$-th layer $Z^{l-1}$, the output of the $l$-th layer $Z^{l}$ can be calculated as follows:
\begin{align}
	Z'_l &= MSA (LN (Z_{l-1})) + Z_{l-1},\\
	Z_l &= MLP (LN (Z'_l)) + Z'_l.
	\label{eq:encoder0}
\end{align}

Specifically, MSA contains $m$ independent self-attention (SA) modules, as well as a re-projection operation W. The inputs of each independent SA are three parts, query ($Q$), key ($K$, and value ($V$). For the specific output of the $(l-1)$-th layer $Z^{l-1}$, the calculation of SA in $l$-th layer $Z^{l}$ can be defined as follows:
\begin{align}
	\begin{split}
		Q = Z_{l-1}W_Q,
		K &= Z_{l-1}W_K, 
		V =Z_{l-1}W_V,\\
		SA (Z_l) &= softmax (\frac{QK^T}{\sqrt{d}}) V,
	\end{split}
	\label{eq:self_attention1}
\end{align}
where $W_Q$, $W_K$, $W_V$ $\in \mathbb{R}^{d\times \frac{d}{m}}$ are three learnable matrices. The $softmax$ function is applied for the input matrix. MLP consists of two fully-connected (FC) layers with one GELU activation function. The first FC layer expends the dimension of embedding from $d$ to $4d$, and the second one compresses the dimension back to $d$. 

Many recent works introduce the transformer into computer vision tasks. DETR \cite{carion2020end} firstly utilizes a CNN backbone to extract the semantic features and then uses the transformer blocks to regress the bounding boxes with category information. ViT \cite{dosovitskiy2021an} regards an image as a sequence of 16$\times$16 words and directly uses the transformer to perform classifications. DeiT \cite{touvron2020deit} uses proper regularization and achieves better performance than ViT using a much smaller dataset. Several pyramid transformers \cite{chu2021twins,wang2021pyramid} are designed for dense prediction tasks. SETR \cite{SETR} regards semantic segmentation as a sequence-to-sequence task with the transformer.

There are two concurrent researches TransCrowd~\cite{liang2021transcrowd} and BCCT \cite{sun2021boosting}, which explore the power of transformers for crowd counting. TransCrowd utilizes a ViT \cite{dosovitskiy2021an} based the transformer decoder with an additional regression token for weakly-supervised crowd counting. BCCT designs a ViT based  model with various attention mechanisms for fully-supervised crowd counting. Our method is a hybrid framework that utilizes a pyramid vision transformer backbone and a simple convolutional decoder head. Our method outperforms these two methods across several benchmarks in both fully-supervised and weakly-supervised settings.


\section{Conclusion}
In this paper,  we propose a simple pipeline for crowd counting under both weakly and fully-supervised settings.  This pipeline contains four components:
a pyramid vision transformer to better capture global context, a feature aggregation module to make full use of information from coarse to fine, a  regression head to provide  multi-scale receptive fields, and tailed loss functions to stabilize the training process.  Without bells and whistles, our method pushes the state-of-the-art further by clear margins across several popular benchmarks. We hope our method can serve as a strong baseline for further research or be ported to other counting tasks.

%

\clearpage


\bibliography{aaai22}

\begin{thebibliography}{50}
\providecommand{\natexlab}[1]{#1}

\bibitem[{Carion et~al.(2020)Carion, Massa, Synnaeve, Usunier, Kirillov, and
  Zagoruyko}]{carion2020end}
Carion, N.; Massa, F.; Synnaeve, G.; Usunier, N.; Kirillov, A.; and Zagoruyko,
  S. 2020.
\newblock End-to-end object detection with transformers.
\newblock In \emph{European Conference on Computer Vision}, 213--229. Springer.

\bibitem[{{Chen}, {Papandreou}, and {Kokkinos}(2018)}]{chen2018deeplab}
{Chen}, L.-C.; {Papandreou}, G.; and {Kokkinos}, I. 2018.
\newblock DeepLab: Semantic Image Segmentation with Deep Convolutional Nets,
  Atrous Convolution, and Fully Connected CRFs.
\newblock \emph{IEEE Transactions on Pattern Analysis and Machine
  Intelligence}, 40(4): 834--848.

\bibitem[{Chu et~al.(2021{\natexlab{a}})Chu, Tian, Wang, Zhang, Ren, Wei, Xia,
  and Shen}]{chu2021twins}
Chu, X.; Tian, Z.; Wang, Y.; Zhang, B.; Ren, H.; Wei, X.; Xia, H.; and Shen, C.
  2021{\natexlab{a}}.
\newblock Twins: Revisiting the Design of Spatial Attention in Vision
  Transformers.
\newblock In \emph{NeurIPS 2021}.

\bibitem[{Chu et~al.(2021{\natexlab{b}})Chu, Tian, Zhang, Wang, Wei, Xia, and
  Shen}]{chu2021conditional}
Chu, X.; Tian, Z.; Zhang, B.; Wang, X.; Wei, X.; Xia, H.; and Shen, C.
  2021{\natexlab{b}}.
\newblock Conditional Positional Encodings for Vision Transformers.
\newblock \emph{arXiv preprint arXiv:2102.10882}.

\bibitem[{Deng et~al.(2009)Deng, Dong, Socher, Li, Li, and
  Fei-Fei}]{deng2009imagenet}
Deng, J.; Dong, W.; Socher, R.; Li, L.-J.; Li, K.; and Fei-Fei, L. 2009.
\newblock Imagenet: A large-scale hierarchical image database.
\newblock In \emph{2009 IEEE conference on computer vision and pattern
  recognition}, 248--255. Ieee.

\bibitem[{Dosovitskiy et~al.(2021)Dosovitskiy, Beyer, Kolesnikov, Weissenborn,
  Zhai, Unterthiner, Dehghani, Minderer, Heigold, Gelly, Uszkoreit, and
  Houlsby}]{dosovitskiy2021an}
Dosovitskiy, A.; Beyer, L.; Kolesnikov, A.; Weissenborn, D.; Zhai, X.;
  Unterthiner, T.; Dehghani, M.; Minderer, M.; Heigold, G.; Gelly, S.;
  Uszkoreit, J.; and Houlsby, N. 2021.
\newblock An Image is Worth 16x16 Words: Transformers for Image Recognition at
  Scale.
\newblock In \emph{International Conference on Learning Representations}.

\bibitem[{{Fang} et~al.(2020){Fang}, {Li}, {Tu}, {Tan}, and
  {Wang}}]{fang2020face}
{Fang}, Y.; {Li}, Y.; {Tu}, X.; {Tan}, T.; and {Wang}, X. 2020.
\newblock Face completion with Hybrid Dilated Convolution.
\newblock \emph{Signal Processing-image Communication}, 80: 115664.

\bibitem[{{Idrees} et~al.(2013){Idrees}, {Saleemi}, {Seibert}, and
  {Shah}}]{idrees2013multi}
{Idrees}, H.; {Saleemi}, I.; {Seibert}, C.; and {Shah}, M. 2013.
\newblock Multi-source Multi-scale Counting in Extremely Dense Crowd Images.
\newblock In \emph{2013 IEEE Conference on Computer Vision and Pattern
  Recognition}, 2547--2554.

\bibitem[{{Idrees} and {Tayyab}(2018)}]{idrees2018composition}
{Idrees}, H.; and {Tayyab}, M. 2018.
\newblock Composition Loss for Counting, Density Map Estimation and
  Localization in Dense Crowds.
\newblock In \emph{Proceedings of the European Conference on Computer Vision
  (ECCV)}.

\bibitem[{{Jiang}, {Zhang}, and {Xu}(2020)}]{jiang2020attention}
{Jiang}, X.; {Zhang}, L.; and {Xu}, M. 2020.
\newblock Attention Scaling for Crowd Counting.
\newblock In \emph{2020 IEEE/CVF Conference on Computer Vision and Pattern
  Recognition (CVPR)}, 4706--4715.

\bibitem[{{Lei} et~al.(2021){Lei}, {Liu}, {Zhang}, and {Liu}}]{lei2021towards}
{Lei}, Y.; {Liu}, Y.; {Zhang}, P.; and {Liu}, L. 2021.
\newblock Towards using count-level weak supervision for crowd counting.
\newblock \emph{Pattern Recognition}, 109: 107616.

\bibitem[{{Li}, {Zhang}, and {Chen}(2018)}]{li2018csrnet}
{Li}, Y.; {Zhang}, X.; and {Chen}, D. 2018.
\newblock CSRNet: Dilated Convolutional Neural Networks for Understanding the
  Highly Congested Scenes.
\newblock In \emph{2018 IEEE/CVF Conference on Computer Vision and Pattern
  Recognition}, 1091--1100.

\bibitem[{{Lian} and {Li}(2019)}]{lian2019density}
{Lian}, D.; and {Li}, J. 2019.
\newblock Density Map Regression Guided Detection Network for RGB-D Crowd
  Counting and Localization.
\newblock In \emph{2019 IEEE/CVF Conference on Computer Vision and Pattern
  Recognition (CVPR)}, 1821--1830.

\bibitem[{{Liang} et~al.(2021){Liang}, {Chen}, {Xu}, {Zhou}, and
  {Bai}}]{liang2021transcrowd}
{Liang}, D.; {Chen}, X.; {Xu}, W.; {Zhou}, Y.; and {Bai}, X. 2021.
\newblock TransCrowd: Weakly-Supervised Crowd Counting with Transformer.
\newblock \emph{arXiv preprint arXiv:2104.09116}.

\bibitem[{{Liu}, {Weng}, and {Mu}(2019)}]{liu2019recurrent}
{Liu}, C.; {Weng}, X.; and {Mu}, Y. 2019.
\newblock Recurrent Attentive Zooming for Joint Crowd Counting and Precise
  Localization.
\newblock In \emph{2019 IEEE/CVF Conference on Computer Vision and Pattern
  Recognition (CVPR)}, 1217--1226.

\bibitem[{{Liu} et~al.(2020){Liu}, {Lu}, {Zou}, {Xiong}, {Cao}, and
  {Shen}}]{liu2020weighing}
{Liu}, L.; {Lu}, H.; {Zou}, H.; {Xiong}, H.; {Cao}, Z.; and {Shen}, C. 2020.
\newblock Weighing Counts: Sequential Crowd Counting by Reinforcement Learning.
\newblock In \emph{European Conference on Computer Vision}, 164--181.

\bibitem[{{Liu} and {Qiu}(2019)}]{liu2019crowd}
{Liu}, L.; and {Qiu}, Z. 2019.
\newblock Crowd Counting With Deep Structured Scale Integration Network.
\newblock In \emph{2019 IEEE/CVF International Conference on Computer Vision
  (ICCV)}, 1774--1783.

\bibitem[{{Liu} et~al.(2019{\natexlab{a}}){Liu}, {Long}, {Zou}, {Niu}, {Pan},
  and {Wu}}]{liu2019adcrowdnet}
{Liu}, N.; {Long}, Y.; {Zou}, C.; {Niu}, Q.; {Pan}, L.; and {Wu}, H.
  2019{\natexlab{a}}.
\newblock ADCrowdNet: An Attention-Injective Deformable Convolutional Network
  for Crowd Understanding.
\newblock In \emph{2019 IEEE/CVF Conference on Computer Vision and Pattern
  Recognition (CVPR)}, 3225--3234.

\bibitem[{{Liu}, {Huang}, and {Wang}(2018)}]{liu2018receptive}
{Liu}, S.; {Huang}, D.; and {Wang}, Y. 2018.
\newblock Receptive Field Block Net for Accurate and Fast Object Detection.
\newblock In \emph{Proceedings of the European Conference on Computer Vision
  (ECCV)}, 404--419.

\bibitem[{{Liu}, {Salzmann}, and {Fua}(2019)}]{liu2019context}
{Liu}, W.; {Salzmann}, M.; and {Fua}, P. 2019.
\newblock Context-Aware Crowd Counting.
\newblock In \emph{2019 IEEE/CVF Conference on Computer Vision and Pattern
  Recognition (CVPR)}, 5099--5108.

\bibitem[{{Liu}, van~de {Weijer}, and {Bagdanov}(2019)}]{liu2019exploiting}
{Liu}, X.; van~de {Weijer}, J.; and {Bagdanov}, A.~D. 2019.
\newblock Exploiting Unlabeled Data in CNNs by Self-Supervised Learning to
  Rank.
\newblock \emph{IEEE Transactions on Pattern Analysis and Machine
  Intelligence}, 41(8): 1862--1878.

\bibitem[{Liu, Yang, and Ding(2020)}]{liu2020adaptive}
Liu, X.; Yang, J.; and Ding, W. 2020.
\newblock Adaptive mixture regression network with local counting map for crowd
  counting.
\newblock \emph{arXiv preprint arXiv:2005.05776}.

\bibitem[{{Liu} et~al.(2019{\natexlab{b}}){Liu}, {Shi}, {Zhao}, and
  {Wang}}]{liu2019point}
{Liu}, Y.; {Shi}, M.; {Zhao}, Q.; and {Wang}, X. 2019{\natexlab{b}}.
\newblock Point in, Box Out: Beyond Counting Persons in Crowds.
\newblock In \emph{2019 IEEE/CVF Conference on Computer Vision and Pattern
  Recognition (CVPR)}, 6469--6478.

\bibitem[{Liu et~al.(2021)Liu, Lin, Cao, Hu, Wei, Zhang, Lin, and
  Guo}]{liu2021swin}
Liu, Z.; Lin, Y.; Cao, Y.; Hu, H.; Wei, Y.; Zhang, Z.; Lin, S.; and Guo, B.
  2021.
\newblock Swin transformer: Hierarchical vision transformer using shifted
  windows.
\newblock \emph{arXiv preprint arXiv:2103.14030}.

\bibitem[{Loshchilov and Hutter(2018)}]{loshchilov2018decoupled}
Loshchilov, I.; and Hutter, F. 2018.
\newblock Decoupled Weight Decay Regularization.
\newblock In \emph{International Conference on Learning Representations}.

\bibitem[{Ma et~al.(2019)Ma, Wei, Hong, and Gong}]{ma2019bayesian}
Ma, Z.; Wei, X.; Hong, X.; and Gong, Y. 2019.
\newblock Bayesian loss for crowd count estimation with point supervision.
\newblock In \emph{ICCV}, 6142--6151.

\bibitem[{Meng et~al.(2021)Meng, Zhang, Zhao, Yang, Qian, Huang, and
  Zheng}]{meng2021spatial}
Meng, Y.; Zhang, H.; Zhao, Y.; Yang, X.; Qian, X.; Huang, X.; and Zheng, Y.
  2021.
\newblock Spatial Uncertainty-Aware Semi-Supervised Crowd Counting.
\newblock In \emph{The IEEE International Conference on Computer Vision
  (ICCV)}.

\bibitem[{{Pham} et~al.(2015){Pham}, {Kozakaya}, {Yamaguchi}, and
  {Okada}}]{pham2015count}
{Pham}, V.-Q.; {Kozakaya}, T.; {Yamaguchi}, O.; and {Okada}, R. 2015.
\newblock COUNT Forest: CO-Voting Uncertain Number of Targets Using Random
  Forest for Crowd Density Estimation.
\newblock In \emph{2015 IEEE International Conference on Computer Vision
  (ICCV)}, 3253--3261.

\bibitem[{{Sam}, {Surya}, and {Babu}(2017)}]{sam2017switching}
{Sam}, D.~B.; {Surya}, S.; and {Babu}, R.~V. 2017.
\newblock Switching Convolutional Neural Network for Crowd Counting.
\newblock In \emph{2017 IEEE Conference on Computer Vision and Pattern
  Recognition (CVPR)}, 4031--4039.

\bibitem[{{Shi} et~al.(2019){Shi}, {Yang}, {Xu}, and
  {Chen}}]{shi2019revisiting}
{Shi}, M.; {Yang}, Z.; {Xu}, C.; and {Chen}, Q. 2019.
\newblock Revisiting Perspective Information for Efficient Crowd Counting.
\newblock In \emph{2019 IEEE/CVF Conference on Computer Vision and Pattern
  Recognition (CVPR)}, 1--10.

\bibitem[{Shi, Mettes, and Snoek(2019)}]{shi2019counting}
Shi, Z.; Mettes, P.; and Snoek, C.~G. 2019.
\newblock Counting with focus for free.
\newblock In \emph{ICCV}, 4200--4209.

\bibitem[{{Sindagi} and {Patel}(2017)}]{sindagi2017generating}
{Sindagi}, V.~A.; and {Patel}, V.~M. 2017.
\newblock Generating High-Quality Crowd Density Maps Using Contextual Pyramid
  CNNs.
\newblock In \emph{2017 IEEE International Conference on Computer Vision
  (ICCV)}, 1879--1888.

\bibitem[{Song et~al.(2021)Song, Wang, Jiang, Wang, Tai, Wang, Li, Huang, and
  Wu}]{song2021rethinking}
Song, Q.; Wang, C.; Jiang, Z.; Wang, Y.; Tai, Y.; Wang, C.; Li, J.; Huang, F.;
  and Wu, Y. 2021.
\newblock Rethinking Counting and Localization in Crowds: A Purely Point-Based
  Framework.
\newblock In \emph{ICCV}.

\bibitem[{Sun et~al.(2021)Sun, Liu, Probst, Paudel, Popovic, and
  Van~Gool}]{sun2021boosting}
Sun, G.; Liu, Y.; Probst, T.; Paudel, D.~P.; Popovic, N.; and Van~Gool, L.
  2021.
\newblock Boosting Crowd Counting with Transformers.
\newblock \emph{arXiv preprint arXiv:2105.10926}.

\bibitem[{{Szegedy} and {Ioffe}(2016)}]{szegedy2016inception}
{Szegedy}, C.; and {Ioffe}, S. 2016.
\newblock Inception-v4, Inception-ResNet and the Impact of Residual Connections
  on Learning.
\newblock In \emph{Proceedings of the Thirty-First AAAI Conference on
  Artificial Intelligence}.

\bibitem[{Touvron et~al.(2020)Touvron, Cord, Douze, Massa, Sablayrolles, and
  J\'egou}]{touvron2020deit}
Touvron, H.; Cord, M.; Douze, M.; Massa, F.; Sablayrolles, A.; and J\'egou, H.
  2020.
\newblock Training data-efficient image transformers \& distillation through
  attention.
\newblock \emph{arXiv preprint arXiv:2012.12877}.

\bibitem[{{Viola} and {Jones}(2001)}]{viola2001robust}
{Viola}, P.; and {Jones}, M. 2001.
\newblock Robust real-time face detection.
\newblock In \emph{Proceedings Eighth IEEE International Conference on Computer
  Vision. ICCV 2001}, volume~57, 137--154.

\bibitem[{{Wan} and {Chan}(2020)}]{wan2020modeling}
{Wan}, J.; and {Chan}, A.~B. 2020.
\newblock Modeling Noisy Annotations for Crowd Counting.
\newblock In \emph{Advances in Neural Information Processing Systems},
  volume~33, 3386--3396.

\bibitem[{Wan, Liu, and Chan(2021)}]{Wan_2021_CVPR}
Wan, J.; Liu, Z.; and Chan, A.~B. 2021.
\newblock A Generalized Loss Function for Crowd Counting and Localization.
\newblock In \emph{Proceedings of the IEEE/CVF Conference on Computer Vision
  and Pattern Recognition (CVPR)}, 1974--1983.

\bibitem[{{Wang} et~al.(2020){Wang}, {Liu}, {Samaras}, and
  {Nguyen}}]{wang2020distribution}
{Wang}, B.; {Liu}, H.; {Samaras}, D.; and {Nguyen}, M.~H. 2020.
\newblock Distribution Matching for Crowd Counting.
\newblock In \emph{Advances in Neural Information Processing Systems},
  volume~33, 1595--1607.

\bibitem[{Wang et~al.(2020)Wang, Gao, Lin, and Li}]{wang2020nwpu}
Wang, Q.; Gao, J.; Lin, W.; and Li, X. 2020.
\newblock NWPU-crowd: A large-scale benchmark for crowd counting and
  localization.
\newblock \emph{IEEE transactions on pattern analysis and machine
  intelligence}, 43(6): 2141--2149.

\bibitem[{{Wang} et~al.(2019){Wang}, {Gao}, {Lin}, and
  {Yuan}}]{wang2019learning}
{Wang}, Q.; {Gao}, J.; {Lin}, W.; and {Yuan}, Y. 2019.
\newblock Learning From Synthetic Data for Crowd Counting in the Wild.
\newblock In \emph{2019 IEEE/CVF Conference on Computer Vision and Pattern
  Recognition (CVPR)}, 8198--8207.

\bibitem[{Wang et~al.(2021)Wang, Xie, Li, Fan, Song, Liang, Lu, Luo, and
  Shao}]{wang2021pyramid}
Wang, W.; Xie, E.; Li, X.; Fan, D.-P.; Song, K.; Liang, D.; Lu, T.; Luo, P.;
  and Shao, L. 2021.
\newblock Pyramid Vision Transformer: A Versatile Backbone for Dense Prediction
  without Convolutions.
\newblock In \emph{IEEE ICCV}.

\bibitem[{Wojek et~al.(2012)Wojek, Dollar, Schiele, and
  Perona}]{2012Pedestrian}
Wojek, C.; Dollar, P.; Schiele, B.; and Perona, P. 2012.
\newblock Pedestrian Detection: An Evaluation of the State of the Art.
\newblock \emph{IEEE Transactions on Pattern Analysis Machine Intelligence},
  34(4): 743.

\bibitem[{Xiong et~al.(2019)Xiong, Lu, Liu, Liu, Cao, and Shen}]{xiong2019open}
Xiong, H.; Lu, H.; Liu, C.; Liu, L.; Cao, Z.; and Shen, C. 2019.
\newblock From open set to closed set: Counting objects by spatial
  divide-and-conquer.
\newblock In \emph{ICCV}, 8362--8371.

\bibitem[{{Yan}, {Yuan}, and {Zuo}(2019)}]{yan2019perspective}
{Yan}, Z.; {Yuan}, Y.; and {Zuo}, W. 2019.
\newblock Perspective-Guided Convolution Networks for Crowd Counting.
\newblock In \emph{2019 IEEE/CVF International Conference on Computer Vision
  (ICCV)}, 952--961.

\bibitem[{{Yang} and {Li}(2020)}]{yang2020reverse}
{Yang}, Y.; and {Li}, G. 2020.
\newblock Reverse Perspective Network for Perspective-Aware Object Counting.
\newblock In \emph{2020 IEEE/CVF Conference on Computer Vision and Pattern
  Recognition (CVPR)}, 4374--4383.

\bibitem[{{Yang} et~al.(2020){Yang}, {Li}, {Wu}, {Su}, {Huang}, and
  {Sebe}}]{yang2020weakly}
{Yang}, Y.; {Li}, G.; {Wu}, Z.; {Su}, L.; {Huang}, Q.; and {Sebe}, N. 2020.
\newblock Weakly-Supervised Crowd Counting Learns from Sorting Rather Than
  Locations.
\newblock In \emph{ECCV (8)}, 1--17.

\bibitem[{{Zhang} et~al.(2016){Zhang}, {Zhou}, {Chen}, {Gao}, and
  {Ma}}]{zhang2016single}
{Zhang}, Y.; {Zhou}, D.; {Chen}, S.; {Gao}, S.; and {Ma}, Y. 2016.
\newblock Single-Image Crowd Counting via Multi-Column Convolutional Neural
  Network.
\newblock In \emph{2016 IEEE Conference on Computer Vision and Pattern
  Recognition (CVPR)}, 589--597.

\bibitem[{Zheng et~al.(2021)Zheng, Lu, Zhao, Zhu, Luo, Wang, Fu, Feng, Xiang,
  Torr, and Zhang}]{SETR}
Zheng, S.; Lu, J.; Zhao, H.; Zhu, X.; Luo, Z.; Wang, Y.; Fu, Y.; Feng, J.;
  Xiang, T.; Torr, P.~H.; and Zhang, L. 2021.
\newblock Rethinking Semantic Segmentation from a Sequence-to-Sequence
  Perspective with Transformers.
\newblock In \emph{CVPR}.

\end{thebibliography}

\clearpage

\appendix
\section{Appendix}
\subsection{Dataset.}
We evaluate our method across five benchmarks, including UCF\_CC\_50~\cite{idrees2013multi}, ShanghaiTech Part A and Part B~\cite{zhang2016single}, UCF\_QNRF~\cite{idrees2018composition}, and NWPU-Crowd~\cite{wang2020nwpu}. These datasets differ in image resolution ratios, quantities, crowding degree, and color spaces. The performance on these benchmarks proves that CCTrans can deal with the crowds well under the different situations.

\paragraph{Training setting and hyper-parameter.}
We use the training settings as described in the main paper. But for ShanghaiTech Part B, we change the OT loss~\cite{wang2020distribution} to the Bayesian~\cite{ma2019bayesian} loss and use the batch size of 16 in two GPUs for more accurate results. Because the latter is more suitable to the scenes with dramatic scale and density variances. For UCF\_QNRF dataset, we also use replace the OT loss with the Bayesian loss but use a single A100 GPU for training. And we limit the longer side of the validation images up to 2400 pixels for this dataset. Note that most of the other methods expand the limitation to 3096 pixels. And all the experiment results are the generally best results on the official validation set after 1500 epochs training, which can be easily reproduced under the same training settings. 
\paragraph{UCF\_CC\_50.} This dataset shows a lot of challenges~\cite{idrees2013multi}. It randomly collects only 50 gray images with  serious perspective distortions from the Internet. There are a total of 63,974 head annotations and the average number per image is 1280. Although this small dataset lacks the training data and color information, CCTrans still has great performance in the crowded scene.
\paragraph{ShanghaiTech Part A.} It has 300 images and 182 images in the training and testing sets, respectively\cite{zhang2016single}. These images are randomly crawled from the Internet. And the number of people in these images varies largely with a wide range. With more training data and color information, our results are improved a lot. 
\paragraph{ShanghaiTech Part B.} It has 400 training images and 316 testing images, which are captured by the surveillance cameras in the streets of Shanghai\cite{zhang2016single}. And these images have dramatic intra-scene scale and density variations of crowds. But CCTrans can capture these variances easily because of the context modeling ability of transformer-based backbone.
\paragraph{UCF\_QNRF.} This dataset contains 1,535 images with a total of 1,251,642 head annotations~\cite{idrees2018composition}. The images are divided into the training set with 1,201 images and the testing set with 334 images, respectively. This dataset has much more annotated heads than currently available crowd datasets, and most of the objects in the picture are small in scale. Though this dataset is more crowded with much more small-scale objects, our MDC clock can use the detail information complemented by PFA to extract the features of small-scale objects well.
\paragraph{NWPU-Crowd.} It is also a recently built large-scale and challenging congested dataset \cite{wang2020nwpu}. It consists of 5109 images crawled from the InterNet, elaborately annotating 2133375 people. And these images have varying degrees of scale, density, and crowd number differences. This dataset is the closest to the real distribution of the crowd. The accurate results on this dataset suggest CCTrans is a good option for real production.

\subsection{Visualization}
\begin{figure*}[h]
		\centering
		\subfigure[Results on UCF\_CC\_50]{
			\resizebox{0.48\textwidth}{!}{
				\includegraphics{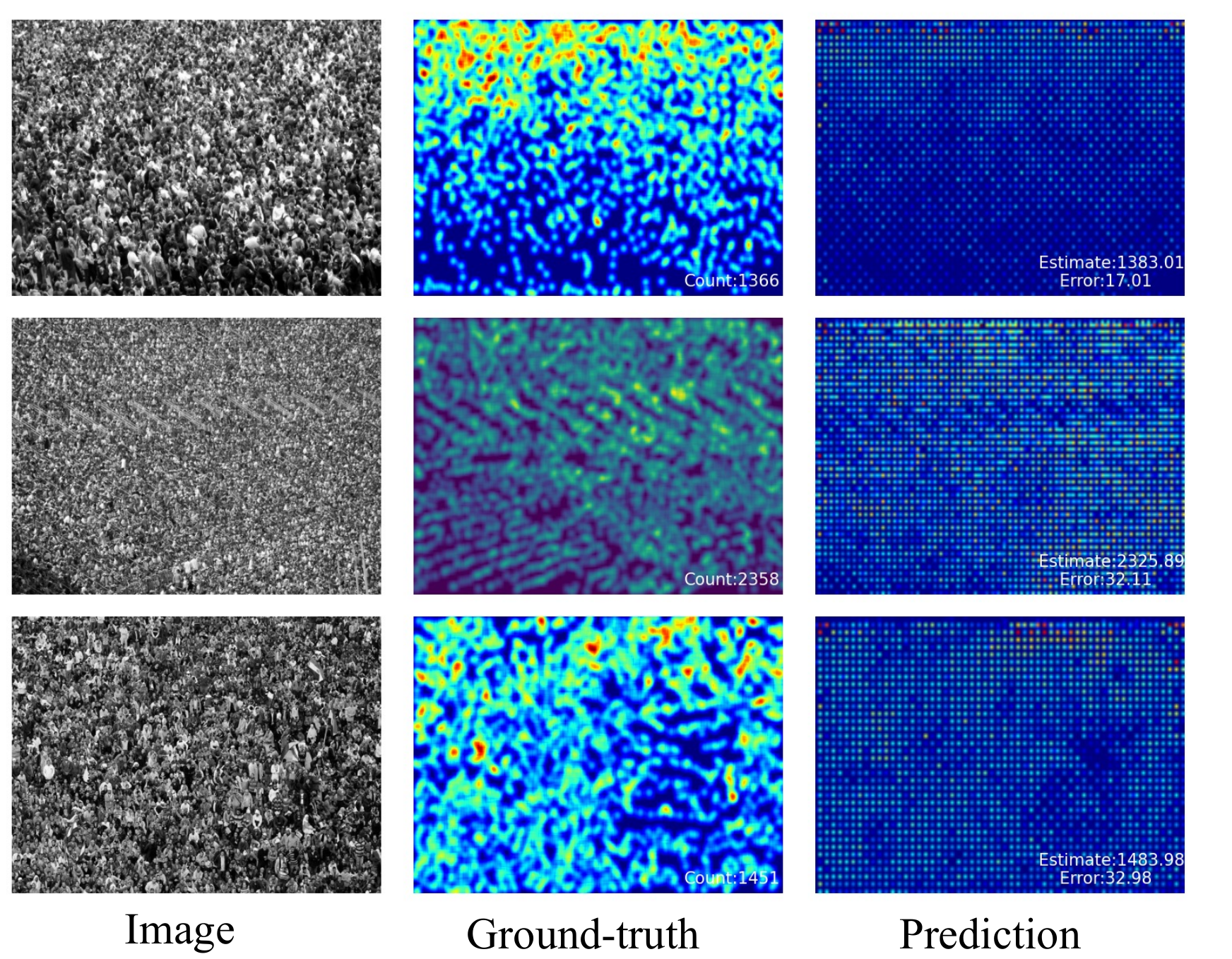}
			}
		}
		\subfigure[Results on ShanghaiTech Part A]{
			\resizebox{0.48\textwidth}{!}{
				\includegraphics{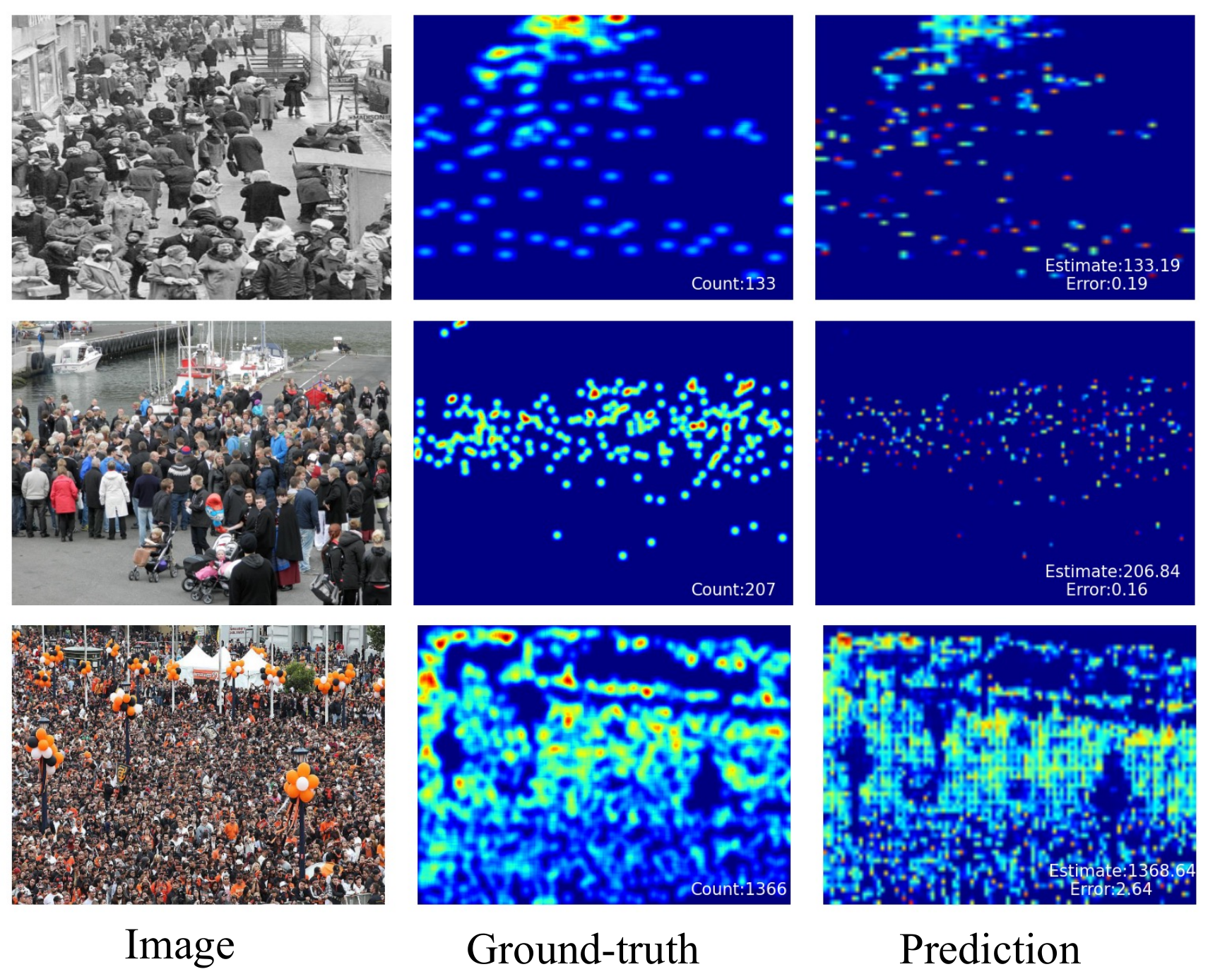}
			}
		}
		
		\subfigure[Results on ShanghaiTech Part B]{
			\resizebox{0.48\textwidth}{!}{
				\includegraphics{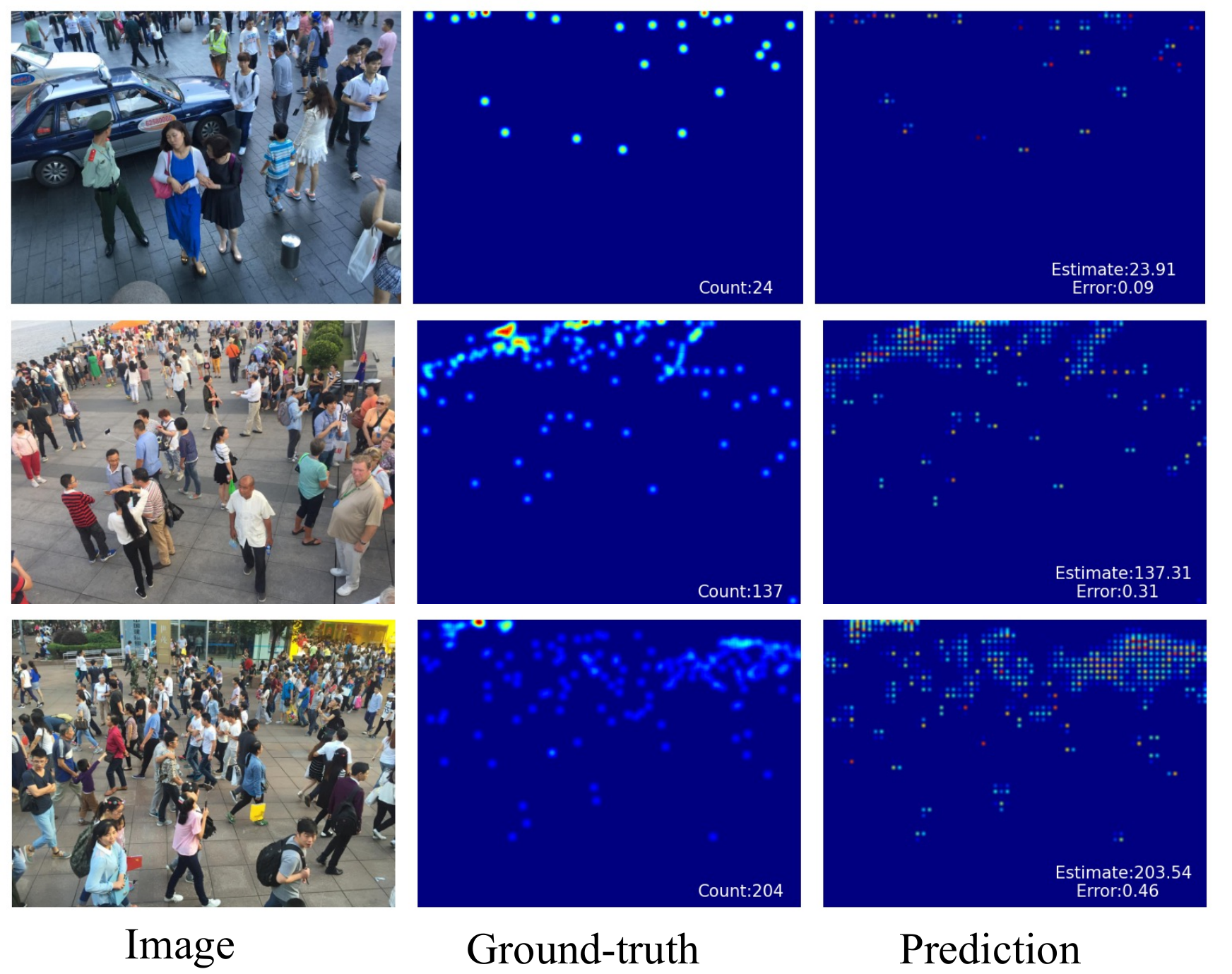}
			}
		}
	
	\centering
	\subfigure[Results on UCF\_QNRF]{
		\resizebox{0.48\textwidth}{!}{
			\includegraphics{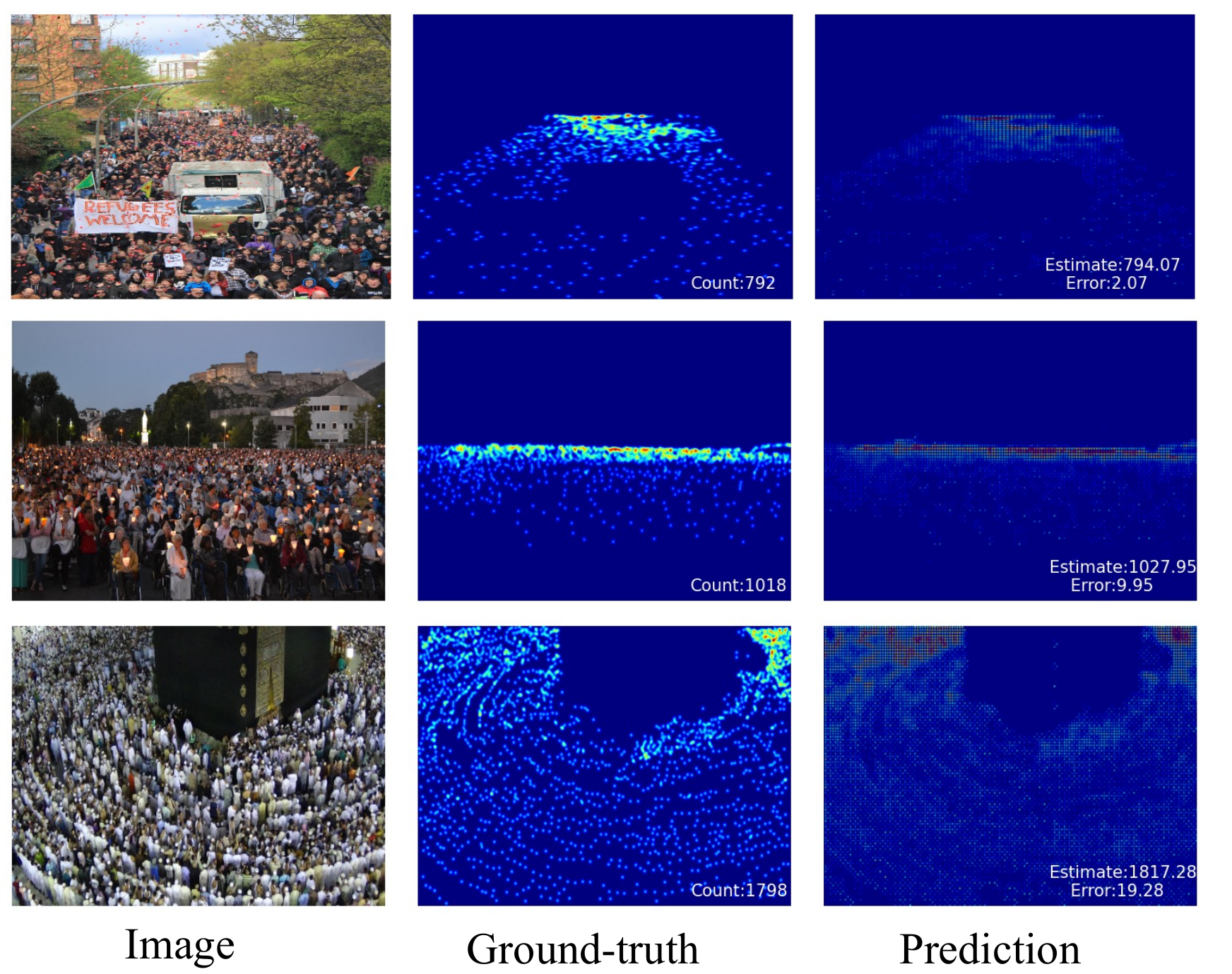}
		}
	}
	\subfigure[Results on NWPU-Crowd]{
		\resizebox{0.48\textwidth}{!}{
			\includegraphics{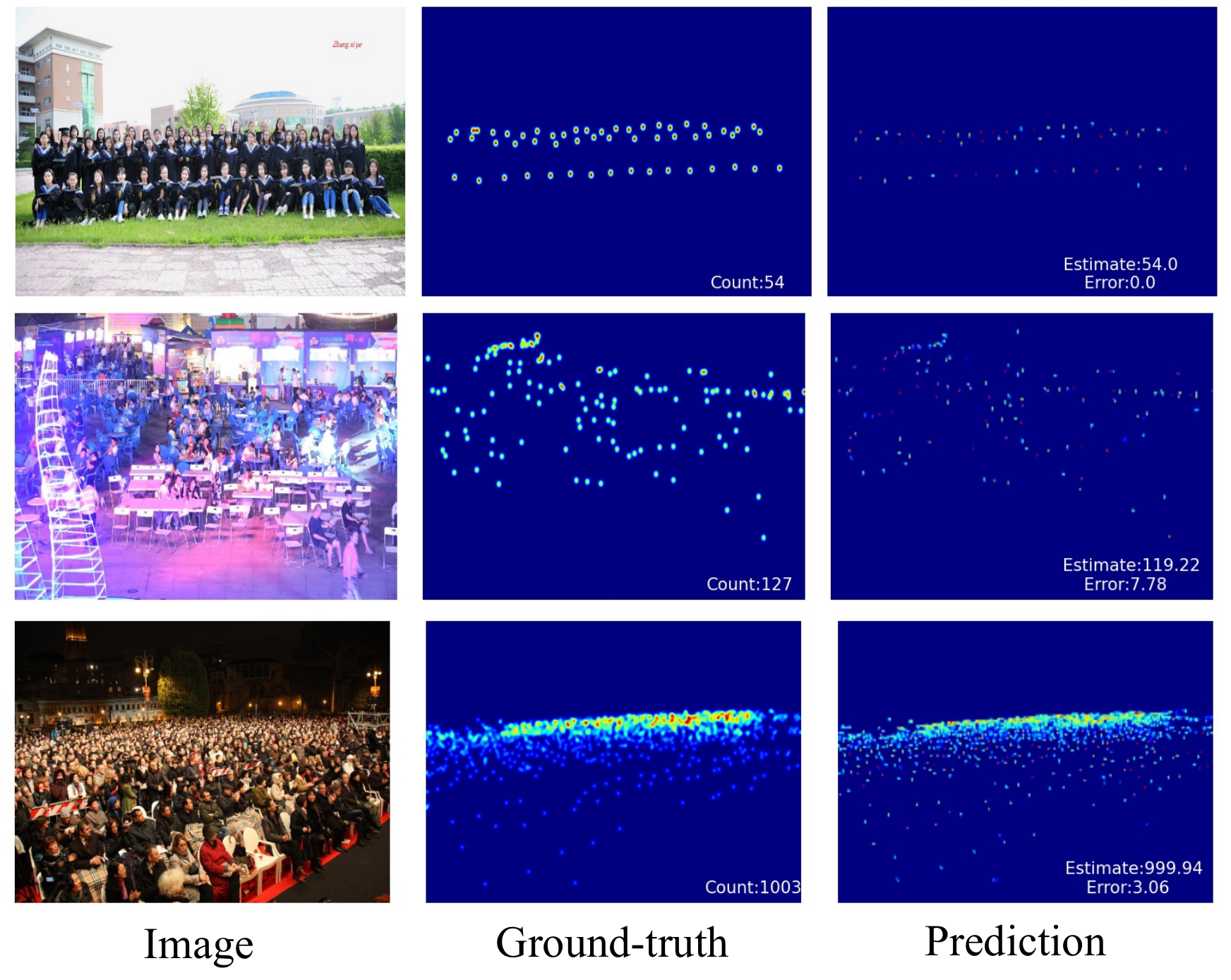}
		}
	}
	\caption{Visualization results of CCTrans on benchmark datasets. The left column shows the original images; the medium column displays the ground-truth density maps while the right column indicates our generated density maps.}
	\label{fig:moreViz}
\end{figure*}

We visualize the density map across three datasets in Fig~\ref{fig:moreViz}. We prove that CCTrans is capable of dealing with images from different sources and colorspaces. And our method also ranks No.1 on the leaderboard of NWPU-Crowd.

\end{document}